\documentclass{ecai}
\usepackage{graphicx}
\usepackage{latexsym}

\usepackage{todonotes}
\usepackage{xcolor}
\usepackage{times}
\usepackage{soul}
\usepackage{url}
\usepackage[hidelinks]{hyperref}
\usepackage[utf8]{inputenc}
\usepackage[small]{caption}
\usepackage{graphicx}
\usepackage{amsmath}
\usepackage{amssymb}
\usepackage{mathtools}
\usepackage{amsthm}
\usepackage{booktabs}
\usepackage{algorithm}
\usepackage{algorithmic}
\usepackage[switch]{lineno}

\usepackage{float}
\usepackage{subfig}

\usepackage{multicol}
\usepackage{nicefrac}       
\usepackage{microtype}      
\usepackage{wrapfig,lipsum,booktabs}
\usepackage{placeins}
\usepackage{multirow}
\usepackage{censor}

\newtheorem{definition}{Definition}
\newtheorem{hypothesis}{Hypothesis}

\theoremstyle{definition}
\newtheorem{example}{Example}

\newenvironment{examplecont}
{\addtocounter{example}{-1}\begin{example}{\textit{\textbf{{(continued)}}}}}
{\end{example}}

\ecaisubmission   

\begin{document}

\begin{frontmatter}

\title{Explaining Image Classification with Visual Debates}

\author[A]{\fnms{Avinash}~\snm{Kori}
\orcid{0000-0002-5878-3584}
\thanks{Corresponding Author. Email: a.kori21@imperial.ac.uk.}}
\author[A]{\fnms{Ben}~\snm{Glocker}
\orcid{0000-0002-4897-9356}
}
\author[A]{\fnms{Francesca}~\snm{Toni}
\orcid{0000-0001-8194-1459}
} 

\address[A]{Department of Computing, Imperial College London}

\begin{abstract}
An effective way to obtain different perspectives on any given topic is by conducting a debate, where participants argue for and against the topic. 
Here, we propose a novel debate framework for understanding and explaining a continuous  image classifier’s reasoning for making a particular prediction by modeling it as a multiplayer sequential zero-sum debate game. 
The contrastive nature of our framework encourages players to learn to put forward diverse arguments during the debates, picking up the reasoning trails missed by their opponents  and highlighting any uncertainties in the classifier. 
Specifically, in our proposed setup,  players propose arguments,
drawn from the classifier’s discretized latent knowledge, to support or oppose the classifier’s decision
. 
The resulting {\em Visual Debates}
collect 
supporting and opposing features from the discretized latent space of the classifier, serving as explanations for the internal reasoning of the classifier towards its predictions.
We  
demonstrate and evaluate (a practical realization of) our  Visual Debates on the geometric SHAPE and MNIST datasets and on the high-resolution animal faces (AFHQ) dataset, along standard  evaluation metrics for explanations (i.e. \emph{faithfulness} and \emph{completeness}) and novel, bespoke metrics for visual debates as explanations (\emph{consensus} and \emph{split ratio}). 
\end{abstract}

\end{frontmatter}

\section{Introduction}
\label{sec:introduction}

Black-box deep learning models can be explained in various ways, 
including feature-attribution \cite{lime,SHAP,deeplift}, attention maps \cite{selvaraju2017grad,gradcampp,integratedcam}, counterfactual explanations \cite{goyal2019counterfactual}, 
neuron level inspection \cite{olah2017feature,olah2020zoom}
or concept based methods \cite{tace,dissect,concepttrail}.
These approaches fall under the category of \emph{post-hoc} explanations, whereby a black-box trained model is diagnosed to extract reasons for making a particular decision.
Instead, methods such as  \cite{rightforright,tellmewhy,debate} aim to develop an 
intrinsically transparent and aligned model.
Each method possesses advantages and disadvantages 
towards
improving the transparency and understanding of the model's reasoning process \cite{doshi2017role,kroll2015accountable}, uncovering model biases \cite{kim2018interpretability},  identifying biases in the data-generating process \cite{narayanaswamy2020scientific}, and 
fulfilling legal obligations for model deployment  \cite{bibal2021legal}.

In the specific
case of 
image classification, most methods providing visual explanations use either heatmaps or localized image segments that are deemed responsible for making a decision \cite{selvaraju2017grad,lime,SHAP,deeplift}.
These explanations capture straightforward input-output relations and do not provide any insights about the model from a debugging or bias mitigation standpoint; 
thus, they fall under the category of \emph{shallow} explanations. 
Another major drawback of these 
methods is that they are a function of the data and the model's prediction
, rather than the model's internal 
states
. Thus, the \emph{faithfulness} of these explanations toward the model's reasoning cannot be 
ascertained. 
Very few current explanation methods follow concept-based reasoning as expressed by humans \cite{armstrong1983some,burnston2021evolving}, with exceptions  \cite{dissect,concepttrail} 
focusing on generating disentangled concepts and traces between them to explain the model's reasoning. 
Recent approaches  \cite{glance,heirarchy} 
keep 
faithfulness in mind and generate explanations using the model's latent knowledge rather than 
the original data
.

We propose a novel method for generating faithful explanations for image classification using \emph{disentangled concepts} obtained by \emph{quantization} \cite{van2017neural}, 
while leveraging on \emph{debates} to trace the model reasoning.
In the case of 
complex data, the discrete, quantized concepts/features in our explanations may not have 
a human understandable meaning, but they nonetheless provide useful information about 
how the model ``reasons''.
Others advocate debate \cite{debate} or dialog \cite{lakkaraju2022rethinking,das2017learning}, 
but not for explanation.
Specifically, 
\cite{debate} advocate debates as an intrinsically transparent and human aligned model demonstrated on a toy setting with the MNIST dataset, while \cite{das2017learning} propose a method to learn natural language dialogs in a cooperative way for reasoning about the considered environment.
Unlike \cite{debate,das2017learning,deng2018visual}, we develop a post-hoc explanation model scalable beyond MNIST, based on non-cooperative interactions between (fictional) players
.
Also, while \cite{lakkaraju2022rethinking} advocate explainability as dialog in principle, we propose a practical framework. 

\begin{figure}[t]
    \centering
    \includegraphics[width=0.48\textwidth]{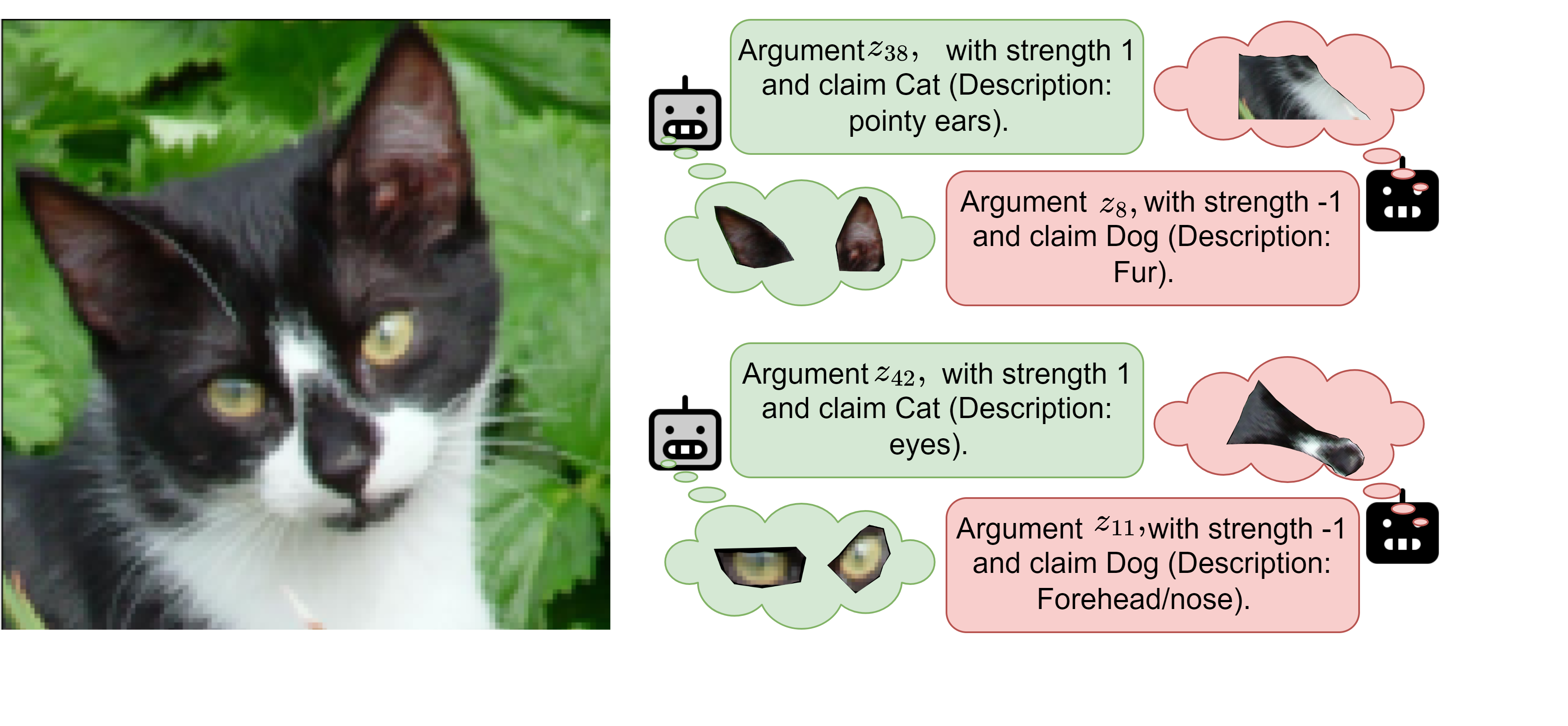}        \caption{Example use of our debate framework with two steps ($n=2$). Here, the classifier $\mathcal{C}$ predicts the input image $x$ as a `Cat', while players $\mathcal{P}^1$ (
    left, grey)  and $\mathcal{P}^2$ (
    right, black) claim the image to be `Cat' and `Dog', respectively. Each argument is 
        characterized by a discrete, quantized feature ($z_l$), a claim (`Cat' or `Dog') and a strength ($\pm 1$).
    (See Example~\ref{example1} and Section~\ref{subsec:debate} for more details.)
    We also include here a human-readable description of the features in the arguments.}
    \label{fig:example}
\end{figure}

We use 
debates as they are influential in bringing out various viewpoints for any given question~\cite{debate}.
For illustration,  consider the question \textit{``Why did the classifier classify this image as a cat?''} (see also Figure~\ref{fig:example}). Debating players may point out quantized features in the image  that are for or against the classifier's decision:
 a player supporting the classifier's decision may pick concepts like \textit{pointy-ears, eyes, or whiskers}, while another player may 
 use other concepts like \textit{fur, forehead or nose} as being characteristic of other animals too, e.g. a dog or a fox. 
This way,
debating  points out, in addition to supporting features, also 
features that oppose the classifier's decision, thus reflecting the classifier's uncertainties. 
While most of the explainable AI methods in the computer vision domain mainly rely on generating single heatmaps or segmenting local image regions responsible for classifiers' decisions \cite{lime, SHAP, selvaraju2017grad, integratedcam, gradcampp, deeplift}, our approach points out both very relevant and possibly ambiguous image regions for the classifiers. 
In general, explanations with relevant reasons and uncertainties help in developing trust of an AI systems~\cite{wang2019deliberative}. 


\begin{example}
 \label{example1}
In Figure \ref{fig:example}, we demonstrate a simple use of our 
method 
with a two step-debate 
between two (fictional) players shown on the right of the figure ($\mathcal{P}^1$ - left, grey -  and $\mathcal{P}^2$ - right, black). 
Here, the classifier predicts the image 
to be of a `Cat' and the players claim  
it to be a `Cat' and a `Dog', respectively. The players' claims are supported by arguments ($z_{38}$,$z_{42}$ for $\mathcal{P}^1$ and
$z_{8}$,$z_{11}$ for $\mathcal{P}^2$)
amounting to quantized features, here visualized as 
regions in the input image and equipped with a human-understandable description. 
%
The debate provides an explanation for the classification, based on the interactions between the players, whereby $\mathcal{P}^1$ makes the first argument ($z_{38}$), indicating  pointy ears as evidence 
for `Cat', rebutted by $\mathcal{P}^2$'s first argument ($z_8$), 
pointing to the fur as evidence against `Cat' and for `Dog', 
 following which $\mathcal{P}^1$ makes the second argument ($z_{42}$), pointing to the eyes as further evidence for `Cat', and thus corroborating $\mathcal{P}^1$'s first argument and rebutting $\mathcal{P}^2$'s argument in turn; finally, $\mathcal{P}^2$ has a shot at rebutting (with $z_{11}$), pointing to the forehead/nose as further evidence for `Dog'. This simple illustration 
 shows 
 how expressive our visual debates can be, unlike conventional explanations such as those in\cite{selvaraju2017grad}, which simply amount to heatmaps, e.g. covering the entire cat (for  examples 
 see \cite{selvaraju2017grad} and Figure~\ref{fig:results}(a) later). 
\end{example}

\begin{figure*}
    \centering
    \includegraphics[width=1.0\textwidth]{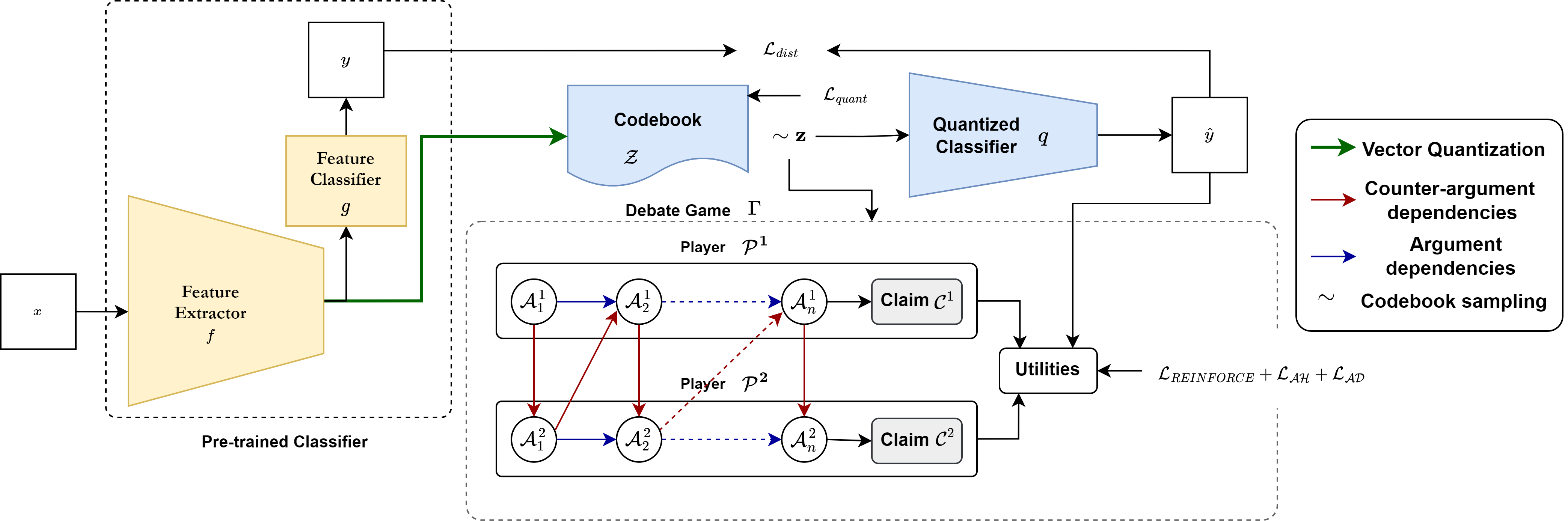}
    \caption{Overview of the proposed realization of the debate framework for classifier $\mathcal{C}=g \circ f$, in which $f$ is the (pre-trained) feature extractor, $g$ is the  (pre-trained) feature classifier, 
    $q$ is the quantized classifier (
    resulting from the `supporting training step', see Section~\ref{subsec:training}) using codebook $\mathcal{Z}$  (
    resulting from the `supporting training step', see Section~\ref{subsec:training})
    for distilling the continuous latent knowledge of $\mathcal{C}$ in a discrete form. $\Gamma$ 
    is a debate game (
    resulting from both the `supportive and contrastive' training steps, see Section~\ref{subsec:training})
    with players $\mathcal{P}^1 \text{ and } \mathcal{P}^2$: $\mathcal{A}^i_k$
    are the arguments put forward by  $\mathcal{P}^i$  at iteration $k$ and  $C^i$ is the claim made by $\mathcal{P}^i$ at the end of the debate
    , when
    each player gets a utility, resulting from argument and claim rewards.   (We restrict attention to two players only for simplicity.)  
    }
    \label{fig:overview}
\end{figure*}

Methodologically, our approach distills the knowledge 
of a trained classifier into a discrete surrogate model (by quantization) and uses the discrete quantized features 
in debates (see Figure~\ref{fig:overview}).
The players involved in the debates learn to pick, from the obtained discrete set of features, the relevant features  to the image as arguments. 
Finally, all the players' arguments 
are used to estimate utilities that 
progressively help 
the players learn. 
The 
arguments in the debate process  are 
visualized by means of a deterministic process 
adapted from \cite{dissection}, 
leading to \emph{visual debates} as explanations (as illustrated in Figure~\ref{fig:example}).
Since we use the latent knowledge from the classifier rather than data
, the generated explanations 
can be deemed to be faithful to the classifier in a sense, as we shall see
.
Overall, our main contributions include:

\begin{itemize}
    \item A \textbf{Debate Framework (Section ~\ref{subsec:debate}
    )} 
    that extracts features 
    for 
    supporting and opposing a continuous image classifier's decision.
    \item \textbf{Player Strategies} and \textbf{Hypothesis Setup (Section ~\ref{subsec:equilibrium})} 
    for analysis 
    at ``equilibrium''.
    \item \textbf{Visual Debates as Explanations (Section~\ref{sec:methods})} and \textbf{their Evaluation (Section~\ref{sec:results})} 
    along four metrics with image classifiers for three datasets (SHAPE \cite{Shapes}, MNIST \cite{deng2012mnist} and AFHQ \cite{choi2020stargan}),  with extensive ablations with respect to debate length and classifiers.
\end{itemize}

\section{Related Work}


Our work is in line with 
\cite{lakkaraju2022rethinking}
, 
advocating the importance of thinking of explainability as a dialog rather than in terms of 
heatmaps or feature attributions, as in much  XAI literature. We give a practical framework in this direction.
Another interesting work which tries to encourage capturing uncertain image regions is \cite{wang2019deliberative}. However, unlike \cite{wang2019deliberative}, we generate both certain and uncertain regions by player interactions in an iterative fashion, while \cite{wang2019deliberative} only focuses on statically capturing ambiguities in an image with respect to given classifier.

We borrow some ideas from 
\cite{debate} to develop our post-hoc explainability framework.
However, 
\cite{debate} focus on 
advocating debate as an ideal candidate for an intrinsically transparent model aligning AI objectives with human values, and demonstrate
a toy 
debate game on the MNIST dataset, 
which proved 
challenging to scale to other natural imaging datasets.
Instead, in the realization of our proposed explanation framework, players are implemented as recurrent attention models \cite{mnih2014recurrent} to ensure generalisability and scalability.

Our proposed debate framework can be seen as generating explanations as ``contests'' between fictional agents (arguing for and against a class). 
There is an emerging interest in developing contestable algorithms, as recent policies for deploying AI systems require the possibility to consider arguments against AI decisions \cite{lyons2021conceptualising,explainableai:Royalsociety}. 
Specifically, \cite{almada2019human,bayamlioglu2018contesting,hirsch2017designing} apply this notion of contestability to sociotechnical systems.
The idea of contestability encourages AI systems to collaborate with humans 
rather than receiving a blind delegation of responsibility. 
This is achieved by providing different avenues of human interactions in the decision-making process \cite{kluttz2022shaping}. 
In our framework, instead, contestability is the very essence of explanation as visual debates. 

 Another 
line of research in XAI involves the use of computational argumentation 
\cite{vcyras2021argumentative}
. 
A common aim for  computational argumentation is to  evaluate a particular claim by considering arguments that support and attack the claim and each other in the context of specific argumentative frameworks (AFs, e.g. of the kinds advocated in \cite{dung1995acceptability} or in \cite{
BAFs}). 
At a high-level of detail, debates can be interpreted as simple instances of AFs, as 
we shall see; however, a fully-fledged analysis of the properties of these AFs is outside the scope of this paper. Instead, we focus on a largely unexplored topic, namely  
explaining image classifiers  via debates as interactive game-playing among 
learning players
.
To the best of our knowledge, our proposed debates are the first to use models' latent knowledge to extract arguments and counterarguments, limiting the action space 
and 
leading to
\emph{Nash Equilibria}. Other approaches using AFs for explainable image classification either use intrinsically argumentative models, e.g. as in \cite{hamed}, or focus on mirroring the mechanics of the model, e.g. as in \cite{Purin21}, rather than explaining with latent features in visual debates.


\section{Preliminaries and Notations}
\label{subsec:prelims}
Let $\mathcal{D} \subseteq \mathcal{X} \times \mathcal{Y}$ be a dataset, such that $\mathcal{X} \in \mathbb{R}^{\tilde{s} \times \tilde{s} \times c}$ and $\mathcal{Y} = \{1, \ldots, N\}$, where $\tilde{s} \times \tilde{}s\times c$ corresponds to the dimension of $\tilde{s} \times \tilde{}s$ images with $c$ channels ($c \geq  1$), and $N\geq 2$ is the total number of classes. 
Let $\mathcal{C}: \mathcal{X} \rightarrow \mathcal{Y}$ be a  {\em classifier} trained on $\mathcal{D}$: 
given an {\em input image} $x \sim \mathcal{X}$, 
$\mathcal{C}(x)\in \mathcal{Y}$ is the predicted class.
As conventional in image classification using deep learning, we assume that $\mathcal{C}$ can be decomposed as $\mathcal{C} = g \circ f$, where $g$ is a \emph{feature classifier} and $f$ is a \emph{feature extractor} (see Figure~\ref{fig:overview}). 

We aim to explain individual predictions by $\mathcal{C}$ in terms of debates whereby (fictional) players exchange arguments, amounting to features supporting classes. 
To achieve this, we assume a model $q$ corresponding  to a \emph{quantized classifier} which behaves like a proxy to the actual feature classifier $g$, where $q: \mathcal{Z} \rightarrow 
\mathbb{P}(\mathcal{Y})$ for $\mathcal{Z}$ a discrete set of \emph{quantized features} \cite{van2017neural} 
and 
$\mathbb{P}(S)$, for any set $S$, is a probability distribution over $S$
, representing the probability for elements of $S$ 
(e.g. determined by a softmax operation). The quantized classifier operates on discrete features rather than continuous features as the original classifier.
The discrete features $z\sim \mathcal{Z}$ are obtained from the input image $x$ as the result of distilling the continuous latent space of the trained classifier, when applied to $x$ (see Figure~\ref{fig:overview}).
For $z \sim \mathcal{Z}, y \in \mathcal{Y}$, we will use $q(z)_y$ to denote the  confidence score for class $y$ as estimated by the quantized classifier $q$ on $z$, and,
with an abuse of notation, we will reserve $q(z)$ to indicate simply the class in $\mathcal{Y}$ with the highest probability/confidence score.
Finally, 
we will use the following notation, 
for $z \sim \mathcal{Z}$, $z' \subseteq z, y \in \mathcal{Y}$:  $q(z; do(z'))_y$ stands for the class confidence score for $y$ estimated by $q$ with just the quantized feature $z'$ and all the other features masked, \emph{i.e.} $\hat{z} = 0, \forall \hat{z} \in z \setminus z'$;
and $q(z; do(z'))$ stands for the class with the highest confidence score after masking. 


\section{Debate Framework}
\label{subsec:debate}
The debate framework consists of 
two \emph{players} $\mathcal{P}^1, \mathcal{P}^2$ who sequentially argue about a common \emph{question} $Q \in \mathcal{Q}$ for a fixed number of $n$ iterations to make \emph{final claims} $\mathcal{C}^1, \mathcal{C}^2$, respectively.\footnote{The framework is actually applicable to any number of players, but we focus for simplicity on two players only. } In the context of explanation for image classification, the question $Q$  relates to the classifier's prediction, and may be something like \textit{``Why did the classifier predict the image $x$ as a cat?''} and the claims may be $\mathcal{C}^1$={\it Cat} and $\mathcal{C}^2$={\it Dog}, as illustrated in Figure~\ref{fig:example}. Player $\mathcal{P}^1$'s objective is to provide relevant \emph{arguments} ($\mathcal{A}^1 = \{\mathcal{A}^1_1, \dots, \mathcal{A}^1_n\}$) supporting the decision made by the classifier on $x$, while player $\mathcal{P}^2$'s objective is to provide relevant counterarguments ($\mathcal{A}^2 = \{\mathcal{A}^2_1, \dots, \mathcal{A}^2_n\}$) 
that oppose the classifier's decision. Here $n$ is the length of/number of steps in the debate: in this paper we assume that this is fixed up-front, depending on the cognitive needs of users using debates as explanations (we experiment with various choices of $n$ in Section~\ref{sec:results}).
Finally, the players are equipped with \emph{utilities}   $\mathcal{U}^1, \mathcal{U}^2$, sanctioning how effective their choices of arguments are towards answering the question with their respective claims. 
Formally, the debate framework is:
$$\Gamma = \langle \{\mathcal{Q}, \mathcal{Z}\}, \{ \mathcal{P}^1, \mathcal{P}^2\},\{\mathcal{A}^1, \mathcal{A}^2\}, \{\mathcal{C}^1, \mathcal{C}^2\},  \{\mathcal{U}^1, \mathcal{U}^2\} \rangle.$$
%
\noindent In  a debate framework, players basically argue about selecting some discrete features for 
supporting ($\mathcal{P}^1$) or for opposing  ($\mathcal{P}^2$)
the classifier's reasoning.
We will define players in Section~\ref{subsec:equilibrium}. 
Here, we define the other components
.

\begin{definition}[Arguments]
\label{def:arg}  An \emph{argument} $\mathcal{A}^i_k$, for $i\in \{1,2\}$ and $k \in \{1, \ldots, n\}$,  is a tuple $(z^i_k, c^i_k, s^i_k)$, where $z^i_k \sim z$ is a particular quantized feature for  $x \sim \mathcal{X}$,
  $c^i_k = \mathrm{argmax}~q(z; do(\{z^i_k\}))$ is the argument \emph{claim}, and
  $s^i_k \in \{-1, 1\}$  is  the argument \emph{strength}
  , where, for 
$\Delta = |q(z)_y - q(z; do(
\{z^1_k,z^2_k\}))_y|$, for $y=\mathcal{C}(x)$ (with $\tau \in (0, 1)$ a given threshold):

\noindent\begin{minipage}{0.49\linewidth}
    \begin{equation*}
        s_k^1 = 
           \begin{cases}
             1,  & \Delta \leq \tau \\
             -1, & \text{otherwise}
           \end{cases}
    \end{equation*}
\end{minipage}
\hfill
\begin{minipage}{0.48\linewidth}
    \begin{equation*}
        s_k^2 = 
           \begin{cases}
             1,  & \Delta > \tau \\
             -1, & \text{otherwise}
           \end{cases}
    \end{equation*}
\end{minipage}

\end{definition}

Here, $\Delta$  measures the effective contribution of a particular pair of 
argument (by $\mathcal{P}^1$) and counterargument (by $\mathcal{P}^2$) towards the quantized classifier's decision.
Note that a low-value of $\Delta$ indicates that the majority of the latent information is encoded in the latent features pair $(z^1_k, z^2_k)$.
Also, the notion of argument strength differs between players: player $\mathcal{P}^1$, supporting the classifier's decision, considers a higher value of $\Delta$ to be a `weak' (negative) argument, while player $\mathcal{P}^2$ considers that to be a `strong' argument, in the spirit of zero-sum 
games.
Further, note that the claim of an argument depends only on the chosen quantized feature in that argument. Also, given $\tau$, the strength of an argument depends only on the quantized feature: thus, we will often equate arguments with their quantized features. 
 


\begin{definition}[Claims]
\label{def:agg} 
%
The \emph{claim} $\mathcal{C}^i$ of  player $\mathcal{P}^i$ is defined as 
$agg^i(\{z^i_1,\ldots, z^i_n\},
\{z^{-i}_1,\ldots, z^{-i}_n\})$, 
for some aggregation function $agg^i: (\cup^{n} \mathcal{Z}) \times (\cup^{n} \mathcal{Z}) \rightarrow \mathcal{Y}$.  
\end{definition}

The notion of player's claim is thus  distinguished from that of argument claim: the former is cumulative and results from the full set of features in arguments (for $\mathcal{P}^1$) and counterarguments (for $\mathcal{P}^2$) at the end of the debate, while the latter only depends on the feature put forward at a particular step.
In practice, in our realization of the debate framework in Section~\ref{sec:methods}, to obtain $agg^i$ we will use hidden state vectors of recurrent neural networks as aggregated arguments, which is followed by a linear layer to map arguments to classes.
Note that different players may perceive the effectiveness of the arguments and counterarguments differently, so different $agg^i$ may result for both players, leading to different players' claims at the end of the debate (after step $n$).

\begin{definition}[Utilities] 
\label{def:util}
Let $Q=q(z,do(\mathcal{A}^1,\mathcal{A}^2))$, and  let 

\( r^i_\Gamma= 
\begin{cases}
1 & \mbox{ if } Q=\mathcal{C}^i, Q\neq \mathcal{C}^{-i}; \\
-1 & \mbox{ if } Q\neq \mathcal{C}^{i}, Q= \mathcal{C}^{-i}; \\
0 & \mbox{ if } Q\neq \mathcal{C}^i, Q\neq \mathcal{C}^{-i} 
\mbox{ or } Q=\mathcal{C}^i, Q= \mathcal{C}^{-i}.
\end{cases}
\)
\\
Then, the \emph{utility} $\mathcal{U}^i$ of player $\mathcal{P}^i$ is 
$\mathcal{U}^i=r^i_\Gamma+\sum_{k=1}^{k=n} s^i_{k}$. 
\end{definition}
Here,  we treat the argument strength $s^i_k$, at step $k$, as an \emph{argument reward}. The utility is a function of the argument rewards and of an overall \emph{claim reward} $r^i_\Gamma$ from the debate,  obtained by masking all the quantized features not present in the arguments 
and comparing the prediction by the quantized classifier after masking ($Q$) with the claims by the players. 
Note that, when  $Q$ matches neither or both claims  by the players, the claim reward is 0, as basically there is no debate between the players in those cases.  
Note also that the utilities, by design, have a zero-sum nature (i.e. $\mathcal{U}^1=-\mathcal{U}^2$), 
reflecting that players should focus on different concepts (quantized features)
.



\begin{examplecont} 
In Figure~\ref{fig:example}, in both steps, the argument rewards are +1 for $\mathcal{P}^1$ and -1  for $\mathcal{P}^2$, resulting in utilities of $\pm 3$, respectively.
The arguments' strength reflects that the classifier is pretty confident that the prediction is correct (as all arguments by $\mathcal{P}^1$ are `strong' and those by $\mathcal{P}^2$ are `weak'),
but in general the argument strength for any player may be any in $\{1,-1\}$.
\end{examplecont}

For debate frameworks to provide \emph{faithful} explanations, 
we need to 
guarantee that the classifier's 
reasoning is encoded in the selected arguments: in Section~\ref{sec:results}, this faithfulness will be measured by computing the accuracy of the debate framework with respect to the classifier's prediction \cite{chen2022makes}.
Given the complexity of  the latent space when classifying images, the discretization by the quantized classifier  helps both in limiting the argument space for the players and, alongside the choice of $n$, in generating \emph{cognitively tractable} explanations~\cite{vcyras2019argumentation}. 

Finally, note that our debates could be interpreted from the perspective of  computational argumentation. For example, each player's argument may be seen as a \emph{rebuttal attack} against each argument by the other player and thus the debate framework may be deemed to form an abstract argumentation framework in the spirit of \cite{dung1995acceptability}, albeit of a restricted form. Moreover, the aggregation function in Definition~\ref{def:agg} could be viewed as a form of \emph{gradual semantics}, in the spirit of \cite{BAFs}.
A re-interpretation of our debates in formal computational argumentation terms and a generalization of the debates to accommodate more complex forms of argumentation are both outside the scope of our paper and are left as interesting directions for future work.

\section{
Players' Strategies and Hypotheses}
\label{subsec:equilibrium}

In this section, we take the individual players' viewpoints and 
formulate hypotheses 
driving our experimental analysis.

So far we have looked at debate frameworks as static objects. 
Here, we demonstrate how players adopt strategies to maximize their utilities for the purpose of generating arguments in debates as faithful explanations for classifiers. 
Players $\mathcal{P}^1$ and $\mathcal{P}^2$ are characterized by parameters $\theta^1$ and $\theta^2$,
respectively, learnable during training (see Section~\ref{sec:methods}).
Strategies amount to (parameterized) probability distributions over choices of quantized features
, conditioned on the quantized features in all  previous arguments (by either players). 

\begin{definition}[
Argument Profiles] 
Players  $\mathcal{P}^i$'s \emph{argument  profile at step} $k \!\in \!\{1, \!\ldots, \! n\}$ are defined as 

$\alpha^1_k\!:\! \bigcup_{t<k}\mathcal{A}^1_t \cup \bigcup_{t<k}\mathcal{A}^2_t %
\rightarrow \mathbb{P}_{\theta^1}
(\mathcal{Z})$ 
and

$\alpha^2_k: 
\bigcup_{t\leq k}\mathcal{A}^1_t \cup \bigcup_{t<k}\mathcal{A}^2_t  \rightarrow \mathbb{P}_{\theta^2}
(\mathcal{Z})
$
. 
\end{definition}
%
Due to the debates' sequential and fully observable nature, an argument made by player $\mathcal {P}^i$ at step $k$ can be used by both players in all the subsequent steps
, and player 2's $k^{th}$ argument is dependent on player 1's $k^{th}$ argument, but not vice versa.
Note that 
players never observe the entire environment at once, but have access to it only via the generate arguments, and the classifier's feedback (reasoning) on the arguments.
In practice, in Section~\ref{sec:methods}, we will enforce (by including them in the objective function during training) properties of \emph{argument entropy minimization}  and \emph{argument diversity maximization}  to 
shape the players' argument profiles, as follows.
\begin{definition}[Argument Entropy ($\mathcal{AH}$)]
For $z \sim \mathcal{Z}$, $\mathcal{AH}
(z) = -\mathbb{E} \log p$, where $p$ 
is the probability of considering a particular feature as an argument. 
\label{dfn:arg_entropy}
\end{definition}
 This notion 
 ensures that the probability distribution over features 
 in an argument 
 profile 
 is focused on a selected few features, 
 leading to more 
 manageable, \emph{cognitively tractable} 
 explanations.

\begin{definition}[Argument Diversity ($\mathcal{AD}$] 
$\mathcal{AD}(\mathcal{A}^1, \mathcal{A}^2) = (\mathbb{E}((\tilde{\mathcal{A}^1}-\mathcal{A}^2)^2) + \mathbb{E}((\mathcal{A}^1 - \tilde{\mathcal{A}^2})^2)) - \lambda \sum_{i \in \{1, 2\}} \mathbb{E}((\mathcal{A}^i - \tilde{\mathcal{A}^i})^2)$, where $\tilde{\mathcal{A}^i} = \frac{1}{n}\sum_{k=1}^{k=n} \mathcal{A}^i_k$ and $\lambda$ is a hyperparameter.
\label{dfn:diversity}
\end{definition}
Intuitively, to preserve the diversity in arguments and encourage coherence between arguments, we maximize the variance between 
inter-player arguments while minimizing the variance between intra-player arguments: the first term in the definition of  $\mathcal{AD}$ measures diversity in arguments and counterarguments made by  the players, while the last term captures the diversity within a player's arguments.

Given an argument profile, a player's \emph{strategy} is any $\mathcal{S}^i_{\theta^i}
= \{z^i_1, \ldots, z^i_n\}$ such that 
$z_k \sim \alpha^i_k(
.)$. 
We will use $\mathbb{P}_{\theta^1}(\mathcal{Z}_k), \mathbb{P}_{\theta^2}(\mathcal{Z}_k)$ to denote the policy
distribution, from which strategies are drawn, for players $\mathcal{P}^1$ and $\mathcal{P}^2$ at step $k$, respectively
. 
%
%
We use $\mathcal{U}^i(\mathcal{S}_{\theta^1}^1,\mathcal{S}_{\theta^2}^2)=r^i_\Gamma+\sum_{t=1}^{t=k} s^i_{t}$ to indicate the utility for a player $\mathcal{P}^i$ 
when both the players use strategies $\mathcal{S}_{\theta^1}^1,\mathcal{S}_{\theta^2}^2$, respectively, within debate framework $\Gamma$ (cf Definition~\ref{def:util} -- as before, $\mathcal{U}^1(\mathcal{S}_{\theta^1}^1,\mathcal{S}_{\theta^2}^2) = -\mathcal{U}^2(\mathcal{S}_{\theta^1}^1,\mathcal{S}_{\theta^2}^2)$). 
Then, players in a debate framework can be seen as participating in a game with joint objective defined in terms of the players' utilities and the log-likelihood of the policy distributions
:
\begin{equation}
    \begin{split}
        V(\mathcal{P}^1
        , \mathcal{P}^2
        ) = &
        \min_{\theta^1} \max_{\theta^2} \mathbb{E} \Big[ \sum_{1\leq k \leq n}  \Big(\log \mathbb{P}_{\theta^1}(\mathcal{Z}_k)
        \\
        & - \log \mathbb{P}_{\theta^2} (\mathcal{Z}_k)\Big)\mathcal{U}^2(\mathcal{S}_{\theta^1}^1, \mathcal{S}_{\theta^2}^2)\Big]
    \end{split}
    \label{eqn:debateobjective}
\end{equation}


This game falls under the category of finite player, finite strategy, fully observable, zero-sum sequential games, which guarantees the existence of at least one mixed strategy \emph{Nash equilibrium (NE)}  \cite{nash1951non} $(\mathcal{S}^i_{\theta^1_*}\mathcal{S}^i_{\theta^2_*})$, where, for player $\mathcal{P}^1$, 
argument $\mathcal{A}^1_k$ made by $\mathcal{P}^{1}$ at step $k>1$ is 
a \emph{best response} to $\mathcal{P}^{2}$'s argument $\mathcal{A}^2_{k-1}$ at the previous step iff
$\mathcal{U}^1(\mathcal{S}^1_{\theta^1_*}, \mathcal{S}^2_{
\theta^2_*}) \geq \mathcal{U}^1(\mathcal{S}^1_{\theta}, \mathcal{S}^2_{
\theta^2_*}) $ for all choices  $\theta$ of parameters (similarly for player $\mathcal{P}^2$). 

When training players for realizing our debate framework (see Section~\ref{sec:methods}),
we aim to make sure that, 
in any NE, both players try to find arguments that align closely with the classifier's reasoning about the input image, with each player trying to uncover the information that the other player failed to capture
.
Thus, our realization of the debate framework is driven by the following hypotheses (assessed empirically in Section~\ref{sec:results}).

\begin{hypothesis}
Both players converge at NE, making true and honest arguments about the 
input image
. 
\label{hyp:debatehypothesis}
\end{hypothesis}
This hypothesis is inspired by
\cite{debate}, who advocate 
it 
for debates for AI safety. In our setting,  the hypothesis is crucial to guarantee that explanations are \emph{faithful} to the classifier's reasoning. 
%

%


\begin{hypothesis}
At any NE, the sampled features $z \sim \mathcal{Z}$ for any 
input image can be 
partitioned into $z_1$ and $z_2$, such that 
$z_1 \cup z_2 = z$ and $z_1 \cap z_2 = \emptyset$, where $z_1$ is a set of features uniquely observed for a given class 
while $z_2$ is a set of features that can be observed for multiple classes. 
At convergence, the arguments made by $\mathcal{P}^1$ and $\mathcal{P}^2$ are sampled from a probability distribution over features $z_1$ and $z_2$ respectively (\emph{i.e.,} $\mathbb{P}_{\theta^1_*}(\mathcal{Z}) \approx \mathbb{P}(z_1), \mathbb{P}_{\theta^2_*}(\mathcal{Z}) \approx \mathbb{P}(z_2)$).
\label{hyp:symbolsplit}
\end{hypothesis}
This hypothesis is plausible due to the design of the argument rewards (as $s^1_k$ can be forced to be 1 if $\mathcal{A}^1_k \in z_1$ and 
 $s^2_k$ can be forced to be 1 if the $\mathcal{A}^2_k \in z_2$
). 
 Features in $z_1$ and $z_2$ correspond, respectively, to semi-factual and counterfactual features as described in \cite{kenny2021generating}, in that, in our debate game, the player $\mathcal{P}^1$, supporting the classifier's class, performs semi-factual perturbations, affecting the internal state of the player while preserving the class, while the other player $\mathcal{P}^2$ performs counter-factual perturbations, affecting the input so as to obtain a different class.
The split of the latent feature space into the two components $z_1$ and $z_2$ is a consequence of these behaviors by the players, which, in turn,
paves the way towards explanations 
that focus on both absolute and uncertain image regions responsible for the classifier's decision making.



\section{Methods}
\label{sec:methods}

In this section, we describe the methods deployed to realize the debate framework and the players behaviour from Sections~\ref{subsec:debate}-\ref{subsec:equilibrium}.  
The proposed realization is 
overviewed in Figure ~\ref{fig:overview}.  

\subsection{Discretization}
\label{subsec:discretization}

The output of $f$ 
is continuous, posing multiple challenges for extracting meaningful explanations about the classifier's latent reasoning.
To address this, we first distill the classifier's latent knowledge into a codebook $\mathcal{Z}$ with $\tilde{n}$ discrete features each of dimension $d$, using the process of vector quantization, similarly to 
\cite{van2017neural}. However, instead of sampling across pixels, we sample along channels, and these channels are disentangled using Hessian penalty of \cite{peebles2020hessian} for obtaining varied features in the codebook. 
We initialize the codebook with a uniform discrete prior, with all $\tilde{n}$ features being uniformly distributed in the range $(-1/\tilde{n}, 1/\tilde{n})$. 
This is done for two main reasons: (i) to bound the vectors with respect to a total number of discrete features, and (ii) to spread all the features within the given range.
The quantization is achieved by deterministically mapping $\tilde{z} = f(x)$ for $x \in \mathcal{X}$ to the nearest codebook vector $z \in \mathcal{Z}$, formally $z = \mathrm{argmin}_j \tilde{d}(\tilde{z}, z_j)$,     for all $ 
z_j\in \mathcal{Z}$ and some convex distance function $\tilde{d}$ (this is captured by$\mathcal{L}_{quant}$ in Figure~\ref{fig:overview}).
In this work we consider two different distance measures: (i) Euclidean sampling: where $\tilde{d}(a, b) = \|a - b \|^2_2$ and (ii) Cosine sampling: where $\tilde{d}(a, b) = -\langle a, b \rangle$.
To distill the knowledge from the continuous to the discrete space, we introduce a quantized classifier $q$, which maps the average pooled sampled vector $z$ to the classifier's decision $\mathcal{C}(x)$ for any given input $x \in \mathcal{X}$.
The parameters in the quantized classifier are trained using knowledge distillation loss $\mathcal{L}_{dist}$, which is the cross-entropy loss between the classifier's decision $\mathcal{C}(x)=y$ and the quantized classifier's prediction $q(z)=\tilde{y}$ (see Figure~\ref{fig:overview}).

\subsection{Players}
\label{subsec:players}

\begin{figure}[!ht]
    \centering
    \includegraphics[width=0.49\textwidth]{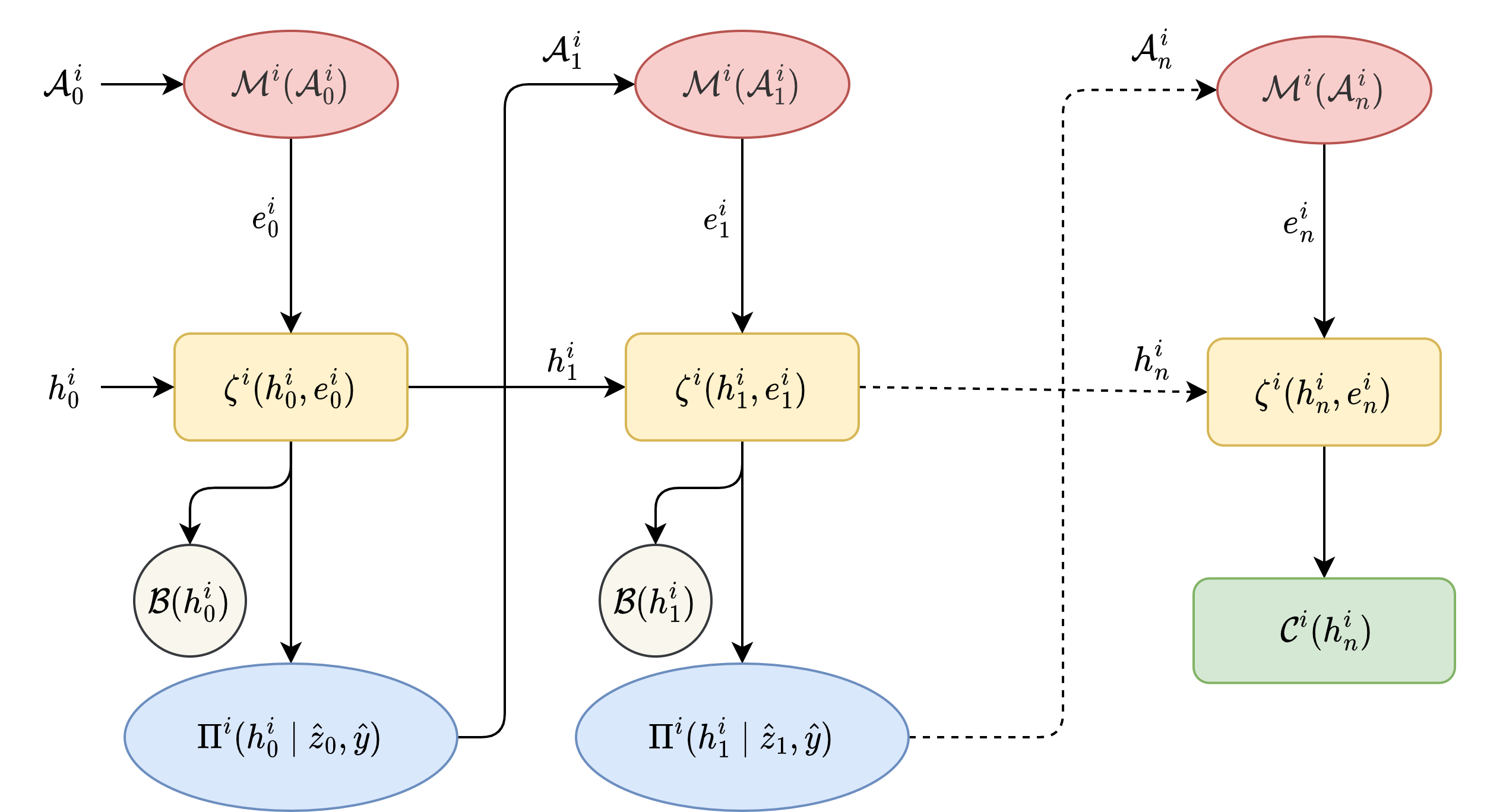}
    \caption{Architecture for Player $\mathcal{P}^i$.
    $\mathcal{M}^i$ is a modulator network and $e^i_k$ is a modulated argument, $\zeta^i$ is 
    a GRU (the backbone network), $h^i_k$ is a hidden state vector
    , $\mathcal{B}^i$, $\Pi^i$ and $\mathcal{C}^i$ are the baseline, policy and claim networks, respectively. $\hat{z}_k$ 
    is a masked state vector.}
    \label{fig:playerFramework}
\end{figure}

To capture the sequential 
nature of debates, we 
implement players with unrolled gated recurrent units (GRU) as the \emph{backbone network $\zeta^i$}.
In addition
, each player also includes a \emph{policy network $\Pi^i$} 
selecting discrete features $z \sim \mathcal{Z}$
 as arguments, a \emph{modulator network $\mathcal{M}^i$} converting arguments to the hidden state dimension in $\zeta^i$ (referred to as modulated arguments), and a \emph{claim network} estimating the player's final 
 claim 
 $\mathcal{C}^i$ (with an abuse of notation, we call this network also $\mathcal{C}^i$).
 We also use a \emph{baseline network $\mathcal{B}^i$} for estimating the value of a particular argument in the spirit of \cite{mnih2014recurrent}.  
Figure \ref{fig:playerFramework} overviews the players' 
architecture and its evolution within a debate with $n$ steps.

At every 
step $k<n$
    (with step 0 some random initialization), $\mathcal{P}^i$ receives an argument $\mathcal{A}^i_{k-1}$ and makes an argument $\mathcal{A}^i_{k}$.
    $\mathcal{A}^i_{k-1}$ is mapped onto representation $e^i_k$ which is used to update the memory of 
    the backbone network $\zeta^i$ to $h^i_{k}$.
Then, the policy network $\Pi^i$ 
considers $h^i_k$ conditioned on 
$\hat{z}_k$ and the quantized classifier's decision $\hat{y}$ as an input to estimate argument $\mathcal{A}^i_{k+1}$, where $\hat{z}$ is generated by masking the environment with respect to $\mathcal{A}^1_{k}$ and $\mathcal{A}^2_{k}$.
At the next step, $\mathcal{M}^i$ transforms $\mathcal{A}^i_{k+1}$ to the required dimension, which is later used in $\zeta^i$ to accumulate the sequential knowledge of arguments made by the players.  
Player $\mathcal{P}^1$ estimates the first argument $\mathcal{A}_1^1$ using a sampled feature and randomly initialized hidden state vector.
Once estimated, arguments are considered common knowledge, and both players can use this information to estimate their subsequent arguments.

\subsection{Training}
\label{subsec:training}

The parameters for a player $\mathcal{P}^i$ can be represented as $\theta^{
i} = \{\theta^i_{\zeta}, \theta^i_{\Pi}, \theta^i_{\mathcal{M}}\}$, where $\theta^i_{\zeta}, \theta^i_{\Pi}$, and  $\theta^i_{\mathcal{M}}$ correspond to the parameters in the 
backbone network $\zeta^i$, policy network $\Pi^i$, and modulator network $\mathcal{M}^i$, respectively. 
We 
use the REINFORCE learning rule with the baseline value to reduce variance in estimates as proposed in \cite{mnih2014recurrent}, averaged over $M$ Monte Carlo samples to update the parameters $\theta^i$:
\begin{equation}
     \frac{1}{M} \sum_m \sum_k \nabla_{\theta^i} \log \Pi^i_{\theta^i}(h^i_k \mid \hat{z}_k, \hat{y}) (\mathcal{U}^i - b^i_k)
    \label{eqn:reinforce}
\end{equation}
%
%
where $\mathcal{U}^i
$ is the player's utility (see Definition~\ref{def:util}) 
and $b^i_k$ corresponds to the baseline value learned by reducing the squared error between $\mathcal{U}^i$ and $b^i_k$.
The gradient in the above learning rule $\nabla_{\theta^i} \log \Pi_{\theta^i}(.)$ can be mapped to gradients of the GRU backbone network 
$\zeta^i$ obtained at step $k$, which can be computed by 
standard backpropagation 
\cite{wierstra2007solving}.
The REINFORCE learning 
rule described above 
allows the player to 
generate an optimal argument sequence based on feedback obtained via delayed cumulative reward ($\mathcal{U}^i$) at the end of every learning episode (i.e. debate of $n$ steps).

The loss term corresponding to the REINFORCE learning rule is described as $\mathcal{L}_{REINFORCE} = -\sum_k \log \Pi^i_{\theta^i}(h^i_k \mid \hat{z}_k, \hat{y}) (\mathcal{U}^i - b^i_k)$.
To train the policy network $\Pi^i$, 
this is combined with argument entropy and argument 
diversity regularisation terms as per Definitions \ref{dfn:arg_entropy} and \ref{dfn:diversity}, represented respectively by 
$\mathcal{L}_{\mathcal{AH}} = \mathcal{AH}(\mathcal{A}^i_k)$ and $\mathcal{L}_{\mathcal{AD}} = -\mathcal{AD}(\mathcal{A}^1, \mathcal{A}^2)$.
These regularisation terms ensure that players make a meaningful argument in any given 
step.
The negative sign in $\mathcal{L}_{\mathcal{AD}}$ and $\mathcal{L}_{\mathcal{AH}}$ is to accommodate 
minimization in the final objective. 

Our training setup consists of two steps: (i) \emph{supportive} training, where both players are trained to support the classifier's decisions, followed by (ii) \emph{contrastive} training, where the players are trained for debating, generating arguments and counterarguments.
In the case of supportive training, we 
use the objective function with utility $\mathcal{U}^1$ for both players along with minimisation of a negative log-likelihood term $\mathcal{L}_{NLL}$ between $\mathcal{C}$'s decision and $\mathcal{P}^1$'s final claim.
Supportive training may be considered a pre-debate step to provide an initial knowledge base for the players to generate arguments and counterarguments in the debate.
The $\mathcal{L}_{NLL}$ term helps to learn better representations in the backbone networks $\zeta^i$, by distilling the classifier's knowledge. 
The 
combined loss during supportive training is  
$\mathcal{L}_{sup.} = \mathcal{L}_{REINFORCE} + \lambda_{1} \mathcal{L}_{NLL} + \lambda_{2} \mathcal{L}_{\mathcal{AH}} - \lambda_{3} \mathcal{L}_{\mathcal{AD}}$. 
The actual debate takes place in the contrastive training step, where players are trained with their respective utilities. The  combined loss during contrastive training is  $\mathcal{L}_{con.} = \mathcal{L}_{REINFORCE} + \lambda_{2} \mathcal{L}_{\mathcal{AH}} + \lambda_{3} \mathcal{L}_{\mathcal{AD}}$.
We do not enforce $\mathcal{L}_{NLL}$ during  contrastive training  to preserve the min-max nature of the debate joint objective in Eq.~\ref{eqn:debateobjective}.

\subsection{
Visualization
}
\label{subsec:explanations}
To visualize arguments in debates, so that they can 
be
comprehensible to human users, 
we follow a deterministic approach based on \cite{dissection}.
We consider, from any specific argument $\mathcal{A}^i_k$,  a low-resolution  attention map $\mathcal{F} \sim z$ 
drawn from the quantized feature $z$ in $\mathcal{A}^i_k$.
We then compare $\mathcal{F}$ with the original input image by resizing it to input-dimension using bilinear interpolation, anchoring the interpolation with respect to the feature's receptive field. 
Specifically, we first normalize the resized attention map between 0 and 1 using minmax normalization 
and overlay it on the original image.
Following this visualization, our debate framework can serve as explanations as \emph{visual debates}, e.g. as illustrated in Figure~\ref{fig:example}.

\begin{table*}[hbt!]
\centering
\caption{Ablation results with respect to debate length (4, 6 or 10) with mean and variance over three random runs.}
\label{table:noncommittedbehavior}

\resizebox{\textwidth}{!}{
\begin{tabular}{@{}ccc|ccc|ccc|ccc|ccc@{}}
\toprule
\multirow{2}{*}{{\begin{tabular}[c]{@{}c@{}}\textsc{Ppty} $\rightarrow$ 
\\ \textsc{Dataset} $\downarrow$\end{tabular}}}  &

\multirow{2}{*}{{\begin{tabular}[c]{@{}c@{}} \textsc{Feature}  
\\ \textsc{Ext.} $\downarrow$\end{tabular}}} &

\multirow{2}{*}{{\begin{tabular}[c]{@{}c@{}}\textsc{Codebook}  
\\ \textsc{Sampling} $\downarrow$\end{tabular}}} &

\multicolumn{3}{c}{\textbf{\begin{tabular}[c]{@{}c@{}}\textsc{Completeness} \end{tabular}}} & 

\multicolumn{3}{c}{\textbf{\begin{tabular}[c]{@{}c@{}} \textsc{Faithfulness} \end{tabular}}} & 

\multicolumn{3}{c}{\textbf{\begin{tabular}[c]{@{}c@{}} \textsc{Consensus} \end{tabular}}} & 

\multicolumn{3}{c}{\textbf{\begin{tabular}[c]{@{}c@{}} \textsc{Split Ratio} ($Z_R$) \end{tabular}}} \\ 

\cmidrule(l){4-15}
&
&
& {4}                     
& {6}                     
& {10}  

& {4}                     
& {6}                     
& {10}  

& {4}                     
& {6}                     
& {10}  

& {4}                     
& {6}                     
& {10}                    \\ \cmidrule(r){1-15}

\multirow{6}{*}{{\begin{tabular}[c]{@{}c@{}} 
\\ \textbf{SHAPE}  \\ \end{tabular}}}

& 
    
\multirow{2}{*}{{\begin{tabular}[c]{@{}c@{}} 
\textsc{Vanilla} \\ ($0.94 \textcolor{gray}{\pm 0.01}$)  \end{tabular}}} 

& \textsc{Euclidian}

& $0.55 \textcolor{gray}{\pm 0.12}$ 
& $0.80 \textcolor{gray}{\pm 0.10}$                             
& $0.85 \textcolor{gray}{\pm 0.02}$

& $0.55 \textcolor{gray}{\pm 0.14}$                             
& $0.79 \textcolor{gray}{\pm 0.07}$                             
& $0.89 \textcolor{gray}{\pm 0.03}$ 

& $0.33 \textcolor{gray}{\pm 0.15}$ 
& $0.64 \textcolor{gray}{\pm 0.06}$                             
& $0.78 \textcolor{gray}{\pm 0.01}$ 

& 0.58                               
& 0.60                               
& 0.60                     \\

&
& \textsc{Cosine}

& $0.58 \textcolor{gray}{\pm 0.23}$                             
& $0.84 \textcolor{gray}{\pm 0.06}$                             
& $0.82  \textcolor{gray}{\pm 0.01}$    

& $0.58 \textcolor{gray}{\pm 0.19}$                             
& $0.81 \textcolor{gray}{\pm 0.03}$                             
& $0.88 \textcolor{gray}{\pm 0.02}$ 

& $0.34 \textcolor{gray}{\pm 0.16}$ 
& $0.76 \textcolor{gray}{\pm 0.07}$                             
& $0.82 \textcolor{gray}{\pm 0.01}$

& 0.48                               
& 0.56                               
& 0.56                     \\





 
\cmidrule(r){2-15}

&
\multirow{2}{*}{{\begin{tabular}[c]{@{}c@{}} 
 \textsc{ResNet}  \\ ($0.98 \textcolor{gray}{\pm 0.01}$) \end{tabular}}} 

& \textsc{Euclidian}

& $0.74 \textcolor{gray}{\pm 0.23}$ 
& $0.83 \textcolor{gray}{\pm 0.14}$                             
& $0.96 \textcolor{gray}{\pm 0.03}$

& $0.76 \textcolor{gray}{\pm 0.12}$                             
& $0.81 \textcolor{gray}{\pm 0.06}$                             
& $0.98 \textcolor{gray}{\pm 0.02}$ 

& $0.41 \textcolor{gray}{\pm 0.12}$ 
& $0.58 \textcolor{gray}{\pm 0.05}$                             
& $0.89 \textcolor{gray}{\pm 0.02}$ 

& 0.38                               
& 0.54                               
& 0.58                     \\

&
& \textsc{Cosine}

& $0.73 \textcolor{gray}{\pm 0.20}$                             
& $0.81 \textcolor{gray}{\pm 0.08}$                             
& $0.97  \textcolor{gray}{\pm 0.02}$    

& $0.76 \textcolor{gray}{\pm 0.21}$                             
& $0.83 \textcolor{gray}{\pm 0.04}$                             
& $0.97 \textcolor{gray}{\pm 0.02}$ 

& $0.44 \textcolor{gray}{\pm 0.09}$ 
& $0.62 \textcolor{gray}{\pm 0.05}$                             
& $0.91 \textcolor{gray}{\pm 0.01}$

& 0.44                               
& 0.58                               
& 0.58                     \\





\cmidrule(r){2-15}

&
\multirow{2}{*}{{\begin{tabular}[c]{@{}c@{}} 
 \textsc{DenseNet} \\ ($0.96 \textcolor{gray}{\pm 0.02}$)  \end{tabular}}} 

& \textsc{Euclidian}

& $0.78 \textcolor{gray}{\pm 0.13}$ 
& $0.89 \textcolor{gray}{\pm 0.04}$                             
& $0.99 \textcolor{gray}{\pm 0.02}$

& $0.81 \textcolor{gray}{\pm 0.07}$                             
& $0.93 \textcolor{gray}{\pm 0.02}$                             
& $0.99 \textcolor{gray}{\pm 0.01}$ 

& $0.63 \textcolor{gray}{\pm 0.07}$ 
& $0.68 \textcolor{gray}{\pm 0.05}$                             
& $0.92 \textcolor{gray}{\pm 0.03}$ 

& 0.36                               
& 0.42                               
& 0.60                     \\

&
& \textsc{Cosine}

& $0.80 \textcolor{gray}{\pm 0.14}$                             
& $0.91 \textcolor{gray}{\pm 0.03}$                             
& $0.98  \textcolor{gray}{\pm 0.02}$    

& $0.83 \textcolor{gray}{\pm 0.13}$                             
& $0.94\textcolor{gray}{\pm 0.06}$                             
& $0.98\textcolor{gray}{\pm 0.01}$ 

& $0.57 \textcolor{gray}{\pm 0.08}$ 
& $0.65 \textcolor{gray}{\pm 0.06}$                             
& $0.90 \textcolor{gray}{\pm 0.03}$

& 0.38                               
& 0.44                               
& 0.58                     \\






\cmidrule(r){1-15}
    
\multirow{6}{*}{{\begin{tabular}[c]{@{}c@{}} 
\\ \textbf{MNIST}  \\ \end{tabular}}}

& 
    
\multirow{2}{*}{{\begin{tabular}[c]{@{}c@{}} 
 \textsc{Vanilla}  \\ ($0.98 \textcolor{gray}{\pm 0.00}$)\end{tabular}}} 

& \textsc{Euclidian}

& $0.36 \textcolor{gray}{\pm 0.22}$ 
& $0.45 \textcolor{gray}{\pm 0.03}$                             
& $0.73 \textcolor{gray}{\pm 0.01}$

& $0.38 \textcolor{gray}{\pm 0.18}$ 
& $0.49 \textcolor{gray}{\pm 0.01}$                             
& $0.83 \textcolor{gray}{\pm 0.00}$ 

& $0.24 \textcolor{gray}{\pm 0.17}$ 
& $0.62 \textcolor{gray}{\pm 0.06}$                             
& $0.74 \textcolor{gray}{\pm 0.04}$ 

& 0.44                               
& 0.45                               
& 0.56     \\

& 
& \textsc{Cosine}

& $0.46 \textcolor{gray}{\pm 0.11}$ 
& $0.58 \textcolor{gray}{\pm 0.06}$                             
& $0.79 \textcolor{gray}{\pm 0.02}$

& $0.53 \textcolor{gray}{\pm 0.08}$                             
& $0.61 \textcolor{gray}{\pm 0.05}$                             
& $0.78 \textcolor{gray}{\pm 0.00}$ 

& $0.22 \textcolor{gray}{\pm 0.11}$ 
& $0.58 \textcolor{gray}{\pm 0.02}$                             
& $0.82 \textcolor{gray}{\pm 0.01}$

& 0.42                               
& 0.45                               
& 0.55     \\






\cmidrule(r){2-15}

& 
    
\multirow{2}{*}{{\begin{tabular}[c]{@{}c@{}} 
 \textsc{ResNet} \\ ($0.99 \textcolor{gray}{\pm 0.00}$) \end{tabular}}} 

& \textsc{Euclidian}

& $0.42 \textcolor{gray}{\pm 0.19}$ 
& $0.46 \textcolor{gray}{\pm 0.08}$                             
& $0.81 \textcolor{gray}{\pm 0.02}$

& $0.45 \textcolor{gray}{\pm 0.13}$ 
& $0.51 \textcolor{gray}{\pm 0.02}$                             
& $0.85 \textcolor{gray}{\pm 0.01}$ 

& $0.18 \textcolor{gray}{\pm 0.21}$ 
& $0.57 \textcolor{gray}{\pm 0.09}$                             
& $0.76 \textcolor{gray}{\pm 0.03}$ 

& 0.54                               
& 0.58                               
& 0.61     \\

& 
& \textsc{Cosine}

& $0.38 \textcolor{gray}{\pm 0.22}$ 
& $0.42 \textcolor{gray}{\pm 0.03}$                             
& $0.78 \textcolor{gray}{\pm 0.00}$

& $0.42 \textcolor{gray}{\pm 0.12}$                             
& $0.49 \textcolor{gray}{\pm 0.05}$                             
& $0.81 \textcolor{gray}{\pm 0.00}$ 

& $0.21 \textcolor{gray}{\pm 0.09}$ 
& $0.58 \textcolor{gray}{\pm 0.02}$                             
& $0.78 \textcolor{gray}{\pm 0.00}$

& 0.43                               
& 0.60                               
& 0.60     \\






\cmidrule(r){2-15}

& 
    
\multirow{2}{*}{{\begin{tabular}[c]{@{}c@{}} 
 \textsc{DenseNet} \\ ($0.99 \textcolor{gray}{\pm 0.00}$) \end{tabular}}} 

& \textsc{Euclidian}                                          

& $0.38 \textcolor{gray}{\pm 0.09}$ 
& $0.54 \textcolor{gray}{\pm 0.05}$                             
& $0.83 \textcolor{gray}{\pm 0.00}$

& $0.41 \textcolor{gray}{\pm 0.15}$ 
& $0.55 \textcolor{gray}{\pm 0.06}$                             
& $0.86 \textcolor{gray}{\pm 0.01}$ 

& $0.31 \textcolor{gray}{\pm 0.21}$ 
& $0.76 \textcolor{gray}{\pm 0.04}$                             
& $0.87 \textcolor{gray}{\pm 0.02}$ 

& 0.39                               
& 0.49                               
& 0.51     \\

& 
& \textsc{Cosine}

& $0.42 \textcolor{gray}{\pm 0.18}$ 
& $0.56 \textcolor{gray}{\pm 0.04}$                             
& $0.77 \textcolor{gray}{\pm 0.01}$

& $0.41 \textcolor{gray}{\pm 0.19}$ 
& $0.56 \textcolor{gray}{\pm 0.09}$                             
& $0.84 \textcolor{gray}{\pm 0.03}$ 

& $0.28 \textcolor{gray}{\pm 0.23}$ 
& $0.74 \textcolor{gray}{\pm 0.06}$                             
& $0.85 \textcolor{gray}{\pm 0.01}$

& 0.44                               
& 0.49                               
& 0.51     \\






\cmidrule(r){1-15}

\multirow{6}{*}{{\begin{tabular}[c]{@{}c@{}} 
\\ \textbf{AFHQ}  \\ \end{tabular}}} 

&
 \multirow{2}{*}{{\begin{tabular}[c]{@{}c@{}} 
 \textsc{Vanilla} \\ $(0.96 \textcolor{gray}{\pm 0.02})$ \end{tabular}}} 

& \textsc{Euclidian}
  
& $0.81 \textcolor{gray}{\pm 0.09}$                            
& $0.86 \textcolor{gray}{\pm 0.06}$                              
& $0.89 \textcolor{gray}{\pm 0.05}$

& $0.89 \textcolor{gray}{\pm 0.02}$                             
& $0.91 \textcolor{gray}{\pm 0.04}$                              
& $0.94 \textcolor{gray}{\pm 0.03}$

& $0.35 \textcolor{gray}{\pm 0.11}$
& $0.43 \textcolor{gray}{\pm 0.08}$  
& $0.82 \textcolor{gray}{\pm 0.04}$  

& 0.39                               
& 0.44                               
& 0.50                     \\ 

&
& \textsc{Cosine}

& $0.31 \textcolor{gray}{\pm 0.10}$                              
& $0.68 \textcolor{gray}{\pm 0.02}$                                
& $0.77 \textcolor{gray}{\pm 0.02}$    

& $0.33 \textcolor{gray}{\pm 0.09}$                                
& $0.71 \textcolor{gray}{\pm 0.04}$                               
& $0.82 \textcolor{gray}{\pm 0.01}$ 

& $0.38 \textcolor{gray}{\pm 0.11}$
& $0.63 \textcolor{gray}{\pm 0.05}$
& $0.76 \textcolor{gray}{\pm 0.02}$

& 0.44                               
& 0.48                               
& 0.49                     \\






\cmidrule(r){2-15}

&
 \multirow{2}{*}{{\begin{tabular}[c]{@{}c@{}} 
 \textsc{ResNet}  \\ ($0.98 \textcolor{gray}{\pm 0.01}$ \end{tabular}}} 

& \textsc{Euclidian}
  
& $0.76 \textcolor{gray}{\pm 0.06}$                            
& $0.84 \textcolor{gray}{\pm 0.01}$                              
& $0.91 \textcolor{gray}{\pm 0.01}$

& $0.81 \textcolor{gray}{\pm 0.07}$                             
& $0.90 \textcolor{gray}{\pm 0.02}$                              
& $0.97 \textcolor{gray}{\pm 0.00}$

& $0.30 \textcolor{gray}{\pm 0.08}$
& $0.69 \textcolor{gray}{\pm 0.03}$
& $0.93 \textcolor{gray}{\pm 0.02}$

& 0.38                               
& 0.41                               
& 0.43                     \\ 

&
& \textsc{Cosine}

& $0.58 \textcolor{gray}{\pm 0.08}$                               
& $0.72 \textcolor{gray}{\pm 0.03}$                                
& $0.88 \textcolor{gray}{\pm 0.00}$    

& $0.64 \textcolor{gray}{\pm 0.06}$                                
& $0.75 \textcolor{gray}{\pm 0.02}$                               
& $0.92 \textcolor{gray}{\pm 0.01}$ 

& $0.40 \textcolor{gray}{\pm 0.18}$
& $0.63 \textcolor{gray}{\pm 0.08}$
& $0.72 \textcolor{gray}{\pm 0.07}$

& 0.40                               
& 0.42                               
& 0.41                     \\






\cmidrule(r){2-15}

&
 \multirow{2}{*}{{\begin{tabular}[c]{@{}c@{}} 
 \textsc{DenseNet} \\ $(0.65 \textcolor{gray}{\pm 0.08})$ \end{tabular}}} 

& \textsc{Euclidian}
  
& $0.49 \textcolor{gray}{\pm 0.12}$                            
& $0.54 \textcolor{gray}{\pm 0.11}$                              
& $0.58 \textcolor{gray}{\pm 0.08}$

& $0.54 \textcolor{gray}{\pm 0.12}$                             
& $0.60 \textcolor{gray}{\pm 0.06}$                              
& $0.62 \textcolor{gray}{\pm 0.02}$

& $0.31 \textcolor{gray}{\pm 0.11}$
& $0.40 \textcolor{gray}{\pm 0.09}$
& $0.56 \textcolor{gray}{\pm 0.04}$

& 0.60                               
& 0.63                               
& 0.68                     \\ 

&
& \textsc{Cosine}

& $0.41 \textcolor{gray}{\pm 0.23}$                               
& $0.40 \textcolor{gray}{\pm 0.08}$                                
& $0.65 \textcolor{gray}{\pm 0.02}$    

& $0.43 \textcolor{gray}{\pm 0.13}$                                
& $0.43 \textcolor{gray}{\pm 0.04}$                               
& $0.65 \textcolor{gray}{\pm 0.03}$ 

& $0.33 \textcolor{gray}{\pm 0.16}$
& $0.48 \textcolor{gray}{\pm 0.02}$
& $0.60 \textcolor{gray}{\pm 0.00}$

& 0.72                               
& 0.72                               
& 0.77                     \\





\bottomrule
\end{tabular}
}
\end{table*}

\subsection{Evaluation Metrics}
\label{sec:eval}
For evaluating the performance of our 
method, we measure and compare properties of \emph{completeness}, \emph{faithfulness}, \emph{consensus}, and \emph{split ratio} across debates of varying debate length, sampling criterion, and feature extractor. 
Formally: (i) \emph{completeness} measures the accuracy of the debate framework with respect to the ground truth labels, by  measuring the data-specific knowledge encoded within the arguments (higher values of completeness 
indicate that meaningful arguments are learned/used in the debates); 
(ii) \emph{faithfulness} measures the accuracy of the debate framework with respect to the classifier's prediction, by quantifying the distillation of the classifier's latent knowledge in the debate framework whereby, as previously discussed, visual debates  are faithful explanations given sufficient arguments (the higher the value of faithfulness the higher the extent of distillation of the classifier's latent knowledge, indicating lower possibilities of debate);
(iii) \emph{consensus} measures how often players agree on the classifier's prediction (given sufficient debate length, the higher the value of consensus 
the lower the ambiguity in the classifier's representations, 
and the lower the consensus the more ambiguous/uncertain the classifier);
and (iv) \emph{split ratio} ($Z_R$) quantitatively measures the existence of splits as per Hypothesis \ref{hyp:symbolsplit}; for $z_1$ and $z_2$ the sets of arguments sampled by $\mathcal{P}^1$ and $\mathcal{P}^2$ respectively for a given input image, $Z_R = \frac{|z_1|}{|z_1| + |z_2|}$, where $|S|$ stands for the cardinality of set $S$.
Thus, the split ratio is the ratio between the total number of uniquely sampled features by $\mathcal{P}^1$ with respect to $\mathcal{P}^2$, and the total number of unique features over the entire training set.
Note that, whereas (i)-(ii) are adaptations to our setting of standard properties in XAI \cite{sokol2020explainability},
(iii)-(iv) are specific to our debates.

\section{Experiments}
\label{sec:results}

We evaluate our debate framework 
with three different CNN classifiers, namely a vanilla network with 5 convolutional layers
, ResNet18~\cite{he2016deep}, and DenseNet121~\cite{huang2017densely}, each of them trained on the SHAPE \cite{Shapes}, MNIST \cite{deng2012mnist} and AFHQ \cite{choi2020stargan} datasets
.
We use images of dimension $32 \times 32
$ with 1 channel for SHAPE and MNIST, and $128 \times 128
$ with 3 channels for AHFQ.
Details about the trained classifiers and training strategies are detailed in the Supplementary Material (SM).
Since, to the best of our knowledge, we are the first to study 
debates for explaining image classifiers, we are not aware of any baseline methods
. 
Thus, to establish players behaviour, measure satisfaction of properties (from Section~\ref{sec:eval}) and verify hypothesis \ref{hyp:symbolsplit}, we conduct extensive ablations on debate length $n$ 
with different codebook samplings (see Section~\ref{subsec:discretization}).

All  findings are in Table \ref{table:noncommittedbehavior}.
The last three columns 
correspond to the split ratio
,  demonstrating that the different players' opinions 
differ about 40-60\% of the time, aligning with hypothesis \ref{hyp:symbolsplit}. Note that difference in arguments does not mean difference in final claims, as the final claims result from the combined effect of all the arguments.
As described in Section \ref{sec:eval}, consensus measures agreement between the players' final claim with respect to the classifier's prediction and is linked to the classifier's uncertainty: based on the results in Table~\ref{table:noncommittedbehavior}, when  increasing debate length, the consensus also tends to increase 
when the model is accurate (a counter-example can be seen for the AFHQ-DenseNet experiment). 
%
Based on the experiments it can also be observed that given sufficient debate length the faithfulness of the generated explanation  also increases, validating our claim 
that visual debates lead to faithful explanations.
Another observation from Table~\ref{table:noncommittedbehavior}
is that, 
for well-performing classifiers 
, the completeness score is similar to the faithfulness score (differing by $\sim 5\%$), again pointing towards faithful explanations.

\begin{figure*}[!ht]
    \centering
    \subfloat[Comparative Explanations]{\includegraphics[width=.45\textwidth]{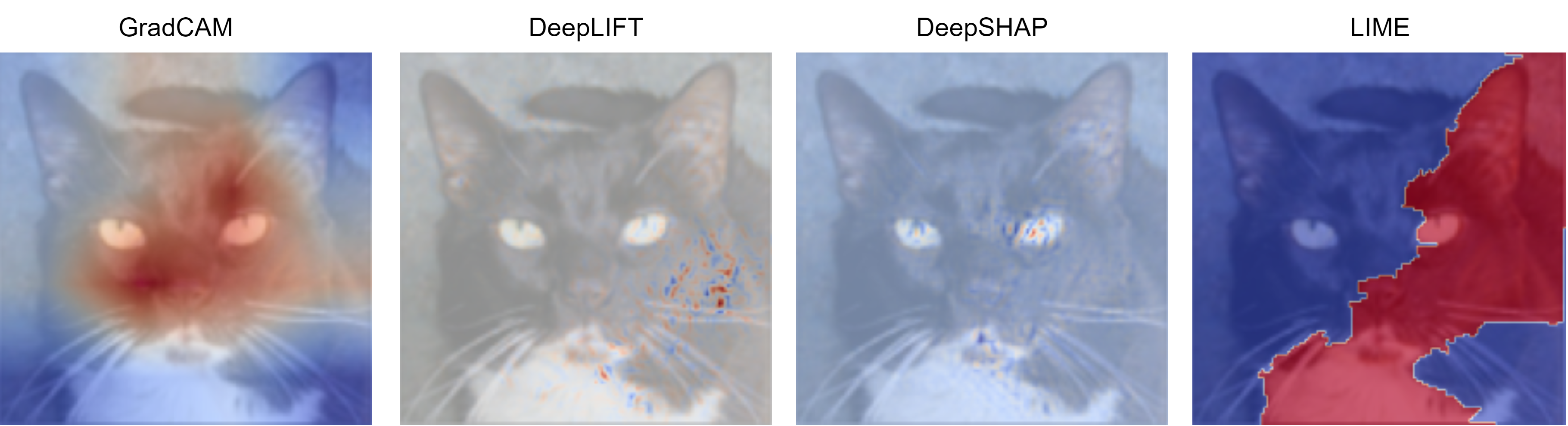}}\hspace{3mm}
    \subfloat[SHAPES dataset, $\mathcal{C}^1=\text{triangle}, \mathcal{C}^2=\text{star}$]{\includegraphics[width=.48\textwidth]{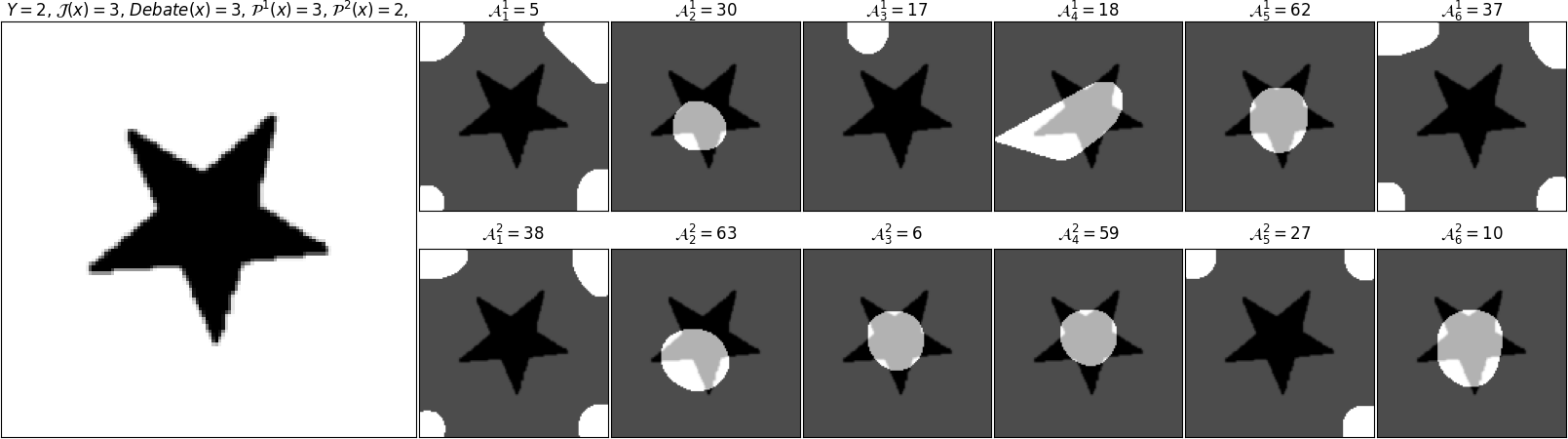}} \hfill
    \subfloat[MNIST dataset, $\mathcal{C}^1=5, \mathcal{C}^2=2$]{\includegraphics[width=.48\textwidth]{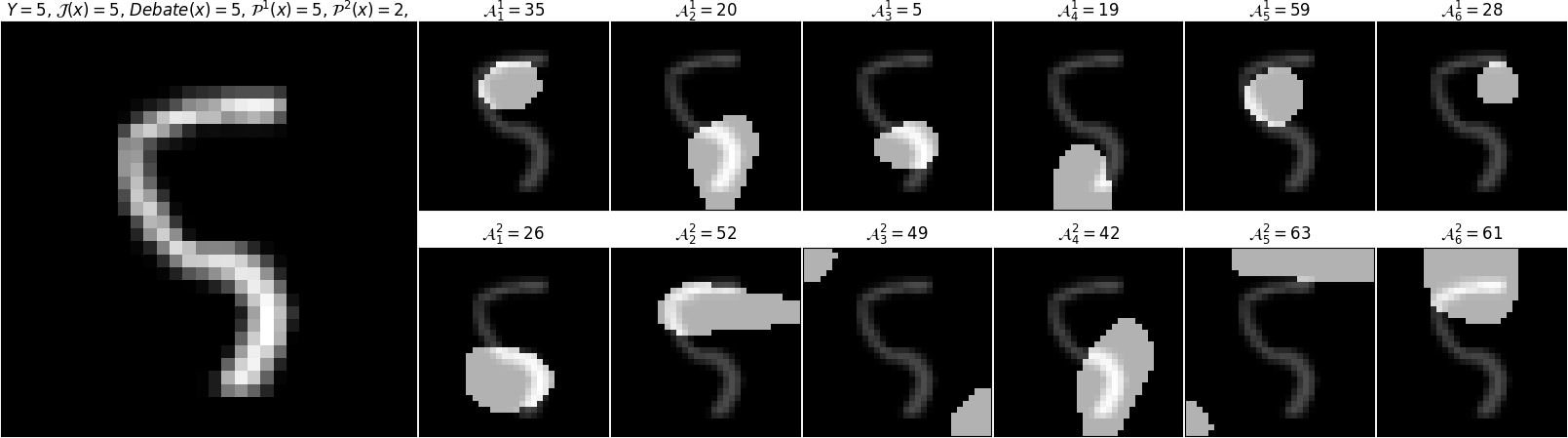}}\hspace{3mm}
    \subfloat[MNIST dataset, $\mathcal{C}^1=8, \mathcal{C}^2=7$]{\includegraphics[width=.48\textwidth]{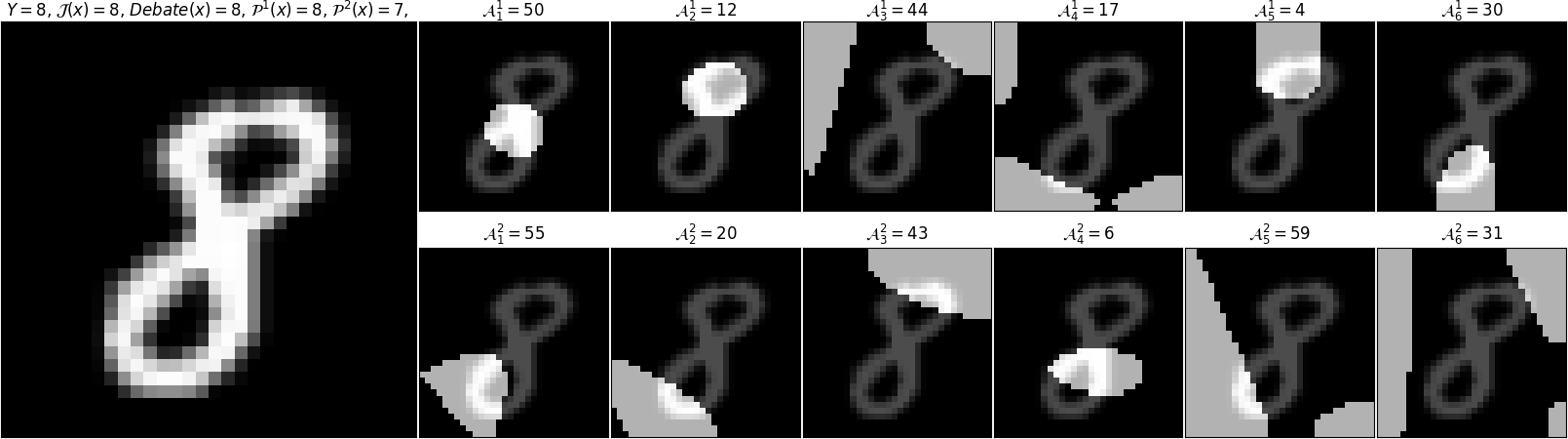}} \hfill    
    \subfloat[AFHQ dataset, $\mathcal{C}^1=\text{cat}, \mathcal{C}^2=\text{wild}$]{\includegraphics[width=.48\textwidth]{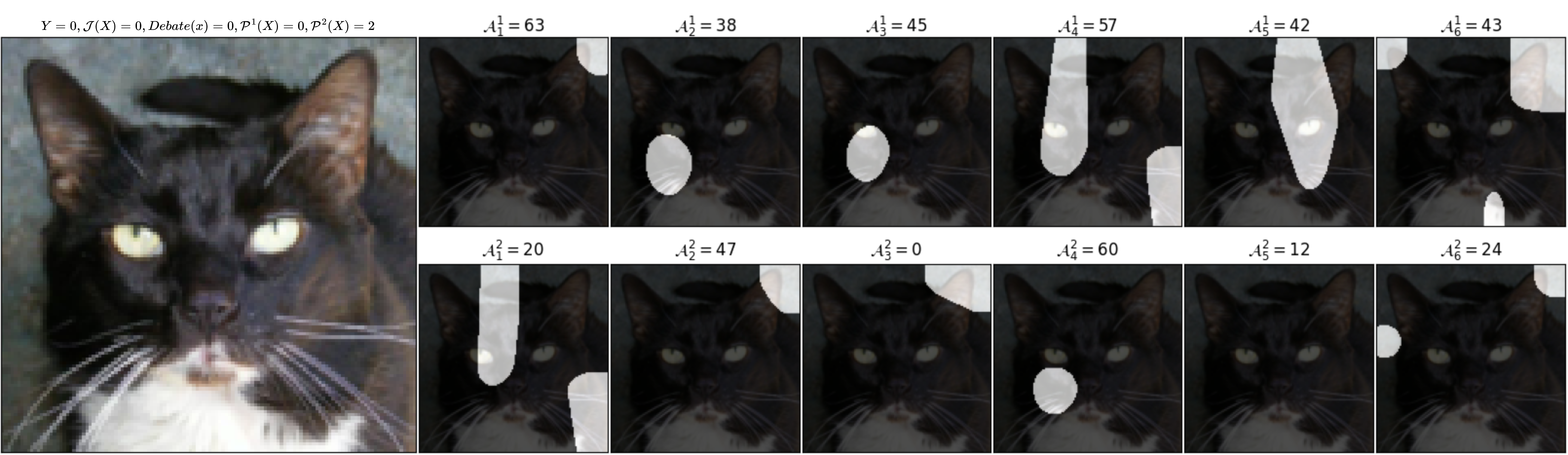}}\hspace{3mm}
    \subfloat[AFHQ dataset, $\mathcal{C}^1=\text{dog}, \mathcal{C}^2=\text{wild}$]{\includegraphics[width=.48\textwidth]{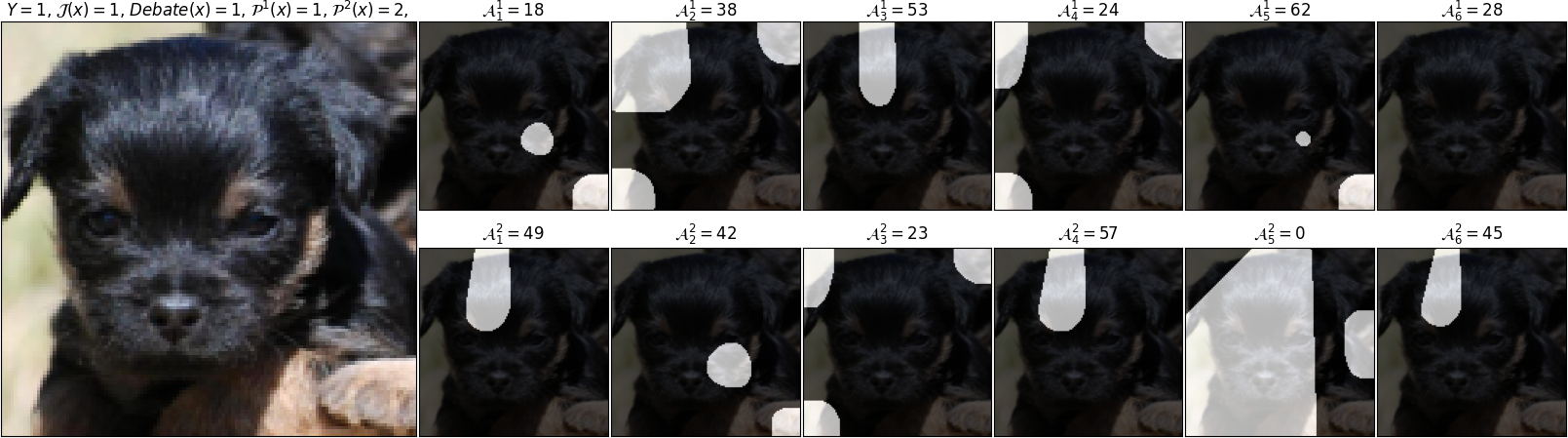}} \hfill
    
    \caption{Explanations obtained by RestNet18 model trained on respective datasets, (a) illustrates explanations obtained from GradCAM, DeepLIFT, DeepSHAP, and LIME from left to right respectively, (b), (c), (d), (e), and (f) illustrates debate explanations. }
    \label{fig:results}
\end{figure*}

To illustrate qualitative difference between the types of explanations obtained from our debate framework and standard feature attribution-based explanations with LIME \cite{lime}, DeepSHAP \cite{SHAP}, deepLIFT \cite{deeplift}, and gradCAM \cite{selvaraju2017grad}, we generated the 
illustrations in Figure \ref{fig:results}.
Figure \ref{fig:results}(a) focuses on existing explanations with AFHQ for comparison, while Figures~\ref{fig:results}(b), \ref{fig:results}(c), \ref{fig:results}(d), \ref{fig:results}(e), and \ref{fig:results}(f) focus on our visual debate explanations on all three datasets.
In the case of the visual debates, the first image is the  input, while the first and second rows correspond respectively to $\mathcal{P}^1$'s and $\mathcal{P}^2$'s visual arguments (their names are their  respective codebook embeddings).
By observing, for instance,  Figures~\ref{fig:results}(a) and \ref{fig:results}(e), it can be seen that the collective debate arguments result in similar explanations as existing methods, but the debate provides additional information by splitting the region of interest into multiple sub-regions as contrasting arguments. 
Thus, the existing methods provide simple input-output explanations (Figure \ref{fig:results}(a)) and disregard the inner reasoning of the classifier, which limits 
their use to debug or reason about the underpinning model.  
Instead, even though the interpretability of the arguments in the visual debates is 
somewhat subjective, 
they help in better pointing out the shortcuts captured by the model.
The illustrations show that, even when the embeddings selected by the players are unique, visual arguments may  overlap, indicating the need for better object centric representations in the classifiers, which we plan to explore and address in future work.


\section{Conclusion}
\label{sec:conclusion}

We defined a novel, practical debate framework for generating post-hoc explanations for image classification as visual debates between (fictional) players. 
We experimentally 
showed  players' convergence 
at Nash Equilibrium and the existence of different sets of features from which players can sample their arguments during debates, confirming our hypotheses. We also showed that our visual debates satisfy desirable properties of completeness, faithfulness, and consensus.


Besides the avenues for future work mentioned in the paper,
we believe that our method could be used to support ontology alignment among humans and models' latent knowledge. 
Our 
debates can also be further explored to develop stand-alone transparent and aligned large scale models. 
Finally, we plan to extend our method with domain experts to assign semantic meaning to arguments and extract a high-level reasoning chain possibly used by classifiers in making 
decisions.


\newpage
\section*{Acknowledgements}
A. Kori was supported by UKRI (grant agreement no. EP/S023356/1), in the UKRI Centre for Doctoral
Training in Safe and Trusted AI.
F. Toni was partially funded
by the European Research Council (ERC) under the European Union’s Horizon 2020 research
and innovation programme (grant agreement No. 101020934) and by J.P. Morgan and the Royal
Academy of Engineering under the Research Chairs and Senior Research Fellowships scheme.
\bibliography{main}

\appendix
\section*{Appendix}

In this appendix, we provide additional details on the players' behavior in our debate framework (Section~\ref{app:obj}), pre-trained classifier details (Section \ref{sec:trainingDetails}), additional ablation experiments and conclusions in section \ref{sec:codebookablation} and finally we illustrate the behavior of committed and non-committed behavior in section \ref{sec:commitnocommit}. We also demonstrate additional debate examples in the end of the SM.

\section{Debate Objective}
\label{app:obj}

As previously described in section 5, the joint objective with respect to the defined policy network $\Pi_{\theta^i}$ in a min-max format can be described as:
\begin{equation}
    \begin{split}
    & V(\mathcal{P}^1_{\theta^1}, \mathcal{P}^2_{\theta^2}) = \min_{\theta^1} \max_{\theta^2} \quad \\
    & \mathbb{E}\Big{[}\sum_t \log \Pi_{\theta^1} (h_t^1 \mid \hat{z}_t, \hat{y})\mathcal{U}^2_t(\mathcal{S}^1, \mathcal{S}^2)\Big{]} \\
    & - \mathbb{E}\Big{[}\sum_t \log \Pi_{\theta^2} (h^2_t\mid  \hat{z}_t, \hat{y})\mathcal{U}^2_t(\mathcal{S}^1, \mathcal{S}^2)\Big{]}
    \end{split}
    \label{eqn:debateobjective1}
\end{equation}

Based on our argument strength and utility definition, we can claim that the utility $\mathcal{U}^1 > 0$ iff the majority of the arguments have positive argument strength.
This ensures that the sampled arguments belong to the semifactual feature set (features, when masked affect class probability but not class outcome).
While for player $\mathcal{P}^2$ we can claim that utility $\mathcal{U}^2 > 0$ the majority of features that are sampled as arguments belong to the counterfactual feature set (features when masked affect the class probability and class outcome).

With this knowledge, if we binarize the utility values, we can restructure the debate objective defined in Equation \ref{eqn:debateobjective1} as follows:

\begin{equation}
    \begin{split}
    & \tilde{V}(\mathcal{P}^1_{\theta^1}, \mathcal{P}^2_{\theta^2}) = 
    \min_{\theta^1} \max_{\theta^2} \quad \\
    & \mathbb{E}\Big{[}\sum_t \log \Pi_{\theta^1} (.)\Big{]} \quad - \mathbb{E}\Big{[}\sum_t \log \Pi_{\theta^2} (.)\Big{]} \\
    & \text{such that: } \mathcal{A}^1_t \in z_1, \mathcal{A}^2_t \in z_2
    \end{split}
    \label{eqn:restructuredobjective}
\end{equation}

Where $z_1$ and $z_2$ correspond to the semi-factual and counter-factual feature sets.
This brings us to our second Hypothesis, which argues about the existence of $z_1$ and $z_2$, such that $z_1, z_2 \subseteq z$ such that $z_1 \cup z_2 = z$ and $z_1 \cap z_2 = \emptyset$ and at convergence forces arguments to follow $\Pi^1_{\theta^1_*}(.) = \mathbb{P}(z_1), \Pi^2_{\theta^2_*}(.) = \mathbb{P}(z_2)$.

\section{Pre-trained models}
\label{sec:trainingDetails}
\textbf{Vanilla model:} We use the custom architecture consisting of 7 convolutional layers with $3 \times 3$ kernel with batch-norm and ReLU activation layer. 
Finally, we project the global average pooled vector onto a class probability space using a linear layer followed by softmax activation. 
To reduce the dimensionality of features, we apply the max pooling layer after the first, third, and fifth layers. 
We train this classifier for 50 epochs with a batch size of 64. We use Adam optimizer with an initial learning rate of 0.001 and weight decay of 0.001. 

\vspace{20pt}
\textbf{Deeper models:} For high-resolution images, we consider the standard DenseNet-121 and Resnet18 architecture and train the model. 
We use Adam optimizer with an initial learning rate of 0.001 and weight decay of 0.005 and trained model for 64 epochs.
\\

\noindent
All our models were trained on a system with GPU: Nvidia Telsa T4 16GB, CPU: Intel(R) Xeon(R) Gold 6230, and RAM of 384GB.

\section{Codebook Ablation and Argument Properties}
\label{sec:codebookablation}
To understand the effect of codebook size on debate accuracy and argument properties, we also tabulate the resulting debate outcome accuracy and split ratio as a result of codebook size and debate length variation.

Table \ref{table:accabl} demonstrates the debate accuracy by varying codebook size and debate length on all three datasets, while Table \ref{table:splitratioabl} demonstrates the variation in split ratio with respect to debate length and codebook size.

Based on this ablation, we claim that:
\begin{itemize}
    \item The debate length helps in achieving better debate accuracy irrespective of codebook size. However, the improvement in performance plateaus after certain length, depending upon dataset. 
    \item Increase in codebook size has an effect on debate performance; we believe this might be because after a certain threshold over codebook size, it makes it easier for players to differentiate between $z_1$ and $z_2$.

\end{itemize}
\begin{table*}
\centering
\caption{Debate faithfulness by varying codebook size (total number of discrete features) on SHAPE, MNIST, and AFHQ datasets.}

\label{table:accabl}
\begin{tabular}{@{}ccccccccccccc@{}}
\toprule
\multicolumn{1}{c}{\textbf{Datasets ($\rightarrow$)}}&
\multicolumn{4}{c}{\textbf{SHAPE}} &
\multicolumn{4}{c}{\textbf{MNIST}} & 
\multicolumn{4}{c}{\textbf{AFHQ}}
\\ \cmidrule(l){2-13} 

\multicolumn{1}{c}{\textbf{Codebook Size in $\mathcal{E}$ ($\downarrow$)}} &

\multicolumn{1}{c}{\textbf{4}} & 
\multicolumn{1}{c}{\textbf{6}} &
\multicolumn{1}{c}{\textbf{10}} &        
\multicolumn{1}{c}{\textbf{20}} &        

\multicolumn{1}{c}{\textbf{4}} & 
\multicolumn{1}{c}{\textbf{6}} &
\multicolumn{1}{c}{\textbf{10}} &        
\multicolumn{1}{c}{\textbf{20}} &        

\multicolumn{1}{c}{\textbf{4}}&         
\multicolumn{1}{c}{\textbf{6}} &
\multicolumn{1}{c}{\textbf{10}} &
\multicolumn{1}{c}{\textbf{20}}
\\ \cmidrule(l){1-13}

\multicolumn{1}{c}{1024} & 

\multicolumn{1}{c}{-} & 
\multicolumn{1}{c}{-} &
\multicolumn{1}{c}{-} & 
\multicolumn{1}{c}{-} & 

\multicolumn{1}{c}{-} & 
\multicolumn{1}{c}{-} &
\multicolumn{1}{c}{-} & 
\multicolumn{1}{c}{-} & 

\multicolumn{1}{c}{{0.61}} &   
\multicolumn{1}{c}{{0.74}} & 
\multicolumn{1}{c}{\textbf{0.83}} & 
\multicolumn{1}{c}{\textbf{0.79}} 

\\

\multicolumn{1}{c}{512} & 

\multicolumn{1}{c}{-} & 
\multicolumn{1}{c}{-} &
\multicolumn{1}{c}{-} & 
\multicolumn{1}{c}{-} & 

\multicolumn{1}{c}{-} & 
\multicolumn{1}{c}{-} &
\multicolumn{1}{c}{-} & 
\multicolumn{1}{c}{-} & 

\multicolumn{1}{c}{{0.63}} &      
\multicolumn{1}{c}{{0.79}} &      
\multicolumn{1}{c}{\textbf{0.83}} &
\multicolumn{1}{c}{\textbf{0.83}} 

\\

\multicolumn{1}{c}{256} & 
\multicolumn{1}{c}{-} & 
\multicolumn{1}{c}{-} &
\multicolumn{1}{c}{-} & 
\multicolumn{1}{c}{-} & 

\multicolumn{1}{c}{-} & 
\multicolumn{1}{c}{-} &
\multicolumn{1}{c}{-} & 
\multicolumn{1}{c}{-} & 

\multicolumn{1}{c}{0.61} &  
\multicolumn{1}{c}{\textbf{0.78}} & 
\multicolumn{1}{c}{\textbf{0.78}} & 
\multicolumn{1}{c}{\textbf{0.82}}

\\

\multicolumn{1}{c}{128} & 

\multicolumn{1}{c}{0.59} & 
\multicolumn{1}{c}{0.82} &
\multicolumn{1}{c}{\textbf{0.93}} &
\multicolumn{1}{c}{\textbf{0.94}} & 

\multicolumn{1}{c}{0.52} & 
\multicolumn{1}{c}{0.64} &
\multicolumn{1}{c}{0.73} &
\multicolumn{1}{c}{\textbf{0.74}} & 

\multicolumn{1}{c}{0.61} &      
\multicolumn{1}{c}{\textbf{0.81}} & 
\multicolumn{1}{c}{\textbf{0.79}} & 
\multicolumn{1}{c}{\textbf{0.81}}

\\

\multicolumn{1}{c}{64} & 

\multicolumn{1}{c}{0.58} & 
\multicolumn{1}{c}{0.80} & 
\multicolumn{1}{c}{0.93} &
\multicolumn{1}{c}{\textbf{0.94}} &

\multicolumn{1}{c}{0.52} & 
\multicolumn{1}{c}{0.64} & 
\multicolumn{1}{c}{0.88} &
\multicolumn{1}{c}{\textbf{0.94}} &

\multicolumn{1}{c}{-} & 
\multicolumn{1}{c}{-} &      
\multicolumn{1}{c}{-} & 
\multicolumn{1}{c}{-}

\\ \bottomrule

\end{tabular}
\end{table*}

\begin{table*}
\centering
\caption{Split ratio by varying codebook size (total number of discrete features) on SHAPE, MNIST, and AFHQ datasets.}

\label{table:splitratioabl}
\begin{tabular}{@{}ccccccccccccc@{}}
\toprule
\multicolumn{1}{c}{\textbf{Datasets ($\rightarrow$)}}&
\multicolumn{4}{c}{\textbf{SHAPE}} &
\multicolumn{4}{c}{\textbf{MNIST}} & 
\multicolumn{4}{c}{\textbf{AFHQ}}
\\ \cmidrule(l){2-13} 

\multicolumn{1}{c}{\textbf{Codebook Size in $\mathcal{E}$ ($\downarrow$)}} &

\multicolumn{1}{c}{\textbf{4}} & 
\multicolumn{1}{c}{\textbf{6}} &
\multicolumn{1}{c}{\textbf{10}} &        
\multicolumn{1}{c}{\textbf{20}} &        

\multicolumn{1}{c}{\textbf{4}} & 
\multicolumn{1}{c}{\textbf{6}} &
\multicolumn{1}{c}{\textbf{10}} &        
\multicolumn{1}{c}{\textbf{20}} &        

\multicolumn{1}{c}{\textbf{4}}&         
\multicolumn{1}{c}{\textbf{6}} &
\multicolumn{1}{c}{\textbf{10}} &
\multicolumn{1}{c}{\textbf{20}}
\\ \cmidrule(l){1-13}

\multicolumn{1}{c}{1024} & 

\multicolumn{1}{c}{-} & 
\multicolumn{1}{c}{-} &
\multicolumn{1}{c}{-} & 
\multicolumn{1}{c}{-} & 

\multicolumn{1}{c}{-} & 
\multicolumn{1}{c}{-} &
\multicolumn{1}{c}{-} & 
\multicolumn{1}{c}{-} & 

\multicolumn{1}{c}{{0.53}} &   
\multicolumn{1}{c}{{0.52}} & 
\multicolumn{1}{c}{\textbf{0.60}} & 
\multicolumn{1}{c}{\textbf{0.60}} 

\\

\multicolumn{1}{c}{512} & 

\multicolumn{1}{c}{-} & 
\multicolumn{1}{c}{-} &
\multicolumn{1}{c}{-} & 
\multicolumn{1}{c}{-} & 

\multicolumn{1}{c}{-} & 
\multicolumn{1}{c}{-} &
\multicolumn{1}{c}{-} & 
\multicolumn{1}{c}{-} & 

\multicolumn{1}{c}{{0.53}} &      
\multicolumn{1}{c}{\textbf{0.58}} &      
\multicolumn{1}{c}{{0.56}} &
\multicolumn{1}{c}{{0.56}} 

\\

\multicolumn{1}{c}{256} & 
\multicolumn{1}{c}{-} & 
\multicolumn{1}{c}{-} &
\multicolumn{1}{c}{-} & 
\multicolumn{1}{c}{-} & 

\multicolumn{1}{c}{-} & 
\multicolumn{1}{c}{-} &
\multicolumn{1}{c}{-} & 
\multicolumn{1}{c}{-} & 

\multicolumn{1}{c}{0.50} &  
\multicolumn{1}{c}{\textbf{0.58}} & 
\multicolumn{1}{c}{\textbf{0.58}} & 
\multicolumn{1}{c}{\textbf{0.58}}

\\

\multicolumn{1}{c}{128} & 

\multicolumn{1}{c}{0.59} & 
\multicolumn{1}{c}{\textbf{0.60}} &
\multicolumn{1}{c}{\textbf{0.58}} &
\multicolumn{1}{c}{\textbf{0.59}} & 

\multicolumn{1}{c}{0.40} & 
\multicolumn{1}{c}{0.47} &
\multicolumn{1}{c}{\textbf{0.58}} &
\multicolumn{1}{c}{\textbf{0.57}} & 

\multicolumn{1}{c}{{0.53}} &      
\multicolumn{1}{c}{\textbf{0.59}} & 
\multicolumn{1}{c}{\textbf{0.59}} & 
\multicolumn{1}{c}{\textbf{0.59}}

\\

\multicolumn{1}{c}{64} & 

\multicolumn{1}{c}{0.58} & 
\multicolumn{1}{c}{0.58} & 
\multicolumn{1}{c}{\textbf{0.60}} &
\multicolumn{1}{c}{\textbf{0.58}} &

\multicolumn{1}{c}{0.38} & 
\multicolumn{1}{c}{0.46} & 
\multicolumn{1}{c}{\textbf{0.56}} &
\multicolumn{1}{c}{\textbf{0.56}} &

\multicolumn{1}{c}{-} & 
\multicolumn{1}{c}{-} &      
\multicolumn{1}{c}{-} & 
\multicolumn{1}{c}{-}

\\ \bottomrule

\end{tabular}
\end{table*}

\begin{table}
\centering
\caption{Ablation results results w.r.t debate length on non-committed games.}
\label{table:noncommittedbehavior}

\begin{tabular}{@{}ccccccc@{}}
\toprule
\multirow{2}{*}{{\begin{tabular}[c]{@{}c@{}}$n$ / $Z_R$ $\rightarrow$ \textbackslash\\ Dataset $\downarrow$\end{tabular}}} &
\multicolumn{3}{c}{\textbf{\begin{tabular}[c]{@{}c@{}}\emph{faithfulness} \end{tabular}}} & 
\multicolumn{3}{c}{\textbf{\begin{tabular}[c]{@{}c@{}}Split Ratio ($Z_R$) \end{tabular}}} \\ \cmidrule(l){2-7} 
& {4}                     
& {6}                     
& {10}                 
& {4}                     
& {6}                     
& {10}                    \\ \cmidrule(r){1-7}

\textbf{SHAPE}                                                                              
& 0.55                               
& 0.79                               
& 0.89                    
& 0.58                               
& 0.60                               
& 0.60                     \\
\textbf{MNIST}                                                                              
& 0.58                               
& 0.61                               
& 0.75                              
& 0.44                               
& 0.45                               
& 0.56                     \\
\textbf{AFHQ}                                                                               
& 0.60                              
& 0.77                               
& 0.80                               
& 0.43                               
& 0.58                               
& 0.54                     \\ 
\bottomrule
\end{tabular}
\end{table}

\section{Committed Vs Non-committed debates}
\label{sec:commitnocommit}
The debates can be categorized into committed and non-committed debates \cite{debate}. 
In the case of committed debates, players are expected to make a claim about the environment at the beginning or at the end of the debate. 
In contrast, in the case of non-committed debates, players argue about the environment without making any claims.
\cite{debate} compare the debate with pre-committed and non-committed behavior and observe that debates with pre-committed claims perform much better in comparison with non-committed debates.
Pre-committed debates have similar properties as non-committed debates, assuming the judge is honest in all the cases.
In \cite{debate}, players reason out all the possibilities without revealing them and make a claim first and provide reasons based on opponents' arguments. 
However, in our setup, as the players are modeled as  POMDP, they reason about an environment only via exchanging arguments and finally use those arguments to make a claim. 
We conduct ablations demonstrating the performance of both committed and non-committed behavior in debate.
In the case of non-committed debates, the claim made by a player is only used in regularisation and not in policy updates, and player $\mathcal{P}^1$ gets a positive reward if the debate outcome is the same as the classifier's decision.

In the case of any NE, given an optimal horizon length $n$ and POMDP structure of players, if arguments and counterarguments made by both players in each and every step are optimal, the debate results in a similar outcome with or without post committing to any claim. We demonstrate this by comparing argument properties and debate convergence via multiple ablations as demonstrated in table \ref{table:noncommittedbehavior} in SM and table 1 in the main text.

\end{document}